\documentclass{article}

\usepackage{arxiv}

\usepackage{algpseudocode}
\usepackage[utf8]{inputenc} % allow utf-8 input
\usepackage[T1]{fontenc}    % use 8-bit T1 fonts
\usepackage{hyperref}       % hyperlinks
\usepackage{url}            % simple URL typesetting
\usepackage{booktabs}       % professional-quality tables
\usepackage{amsfonts}       % blackboard math symbols
\usepackage{nicefrac}       % compact symbols for 1/2, etc.
\usepackage{microtype}      % microtypography
\usepackage{lipsum}		% Can be removed after putting your text content
\usepackage{graphicx}
\usepackage{natbib}
\usepackage{doi}
\usepackage{subcaption}
\usepackage{epsfig}
\usepackage[figuresright]{rotating}
\usepackage{times}

\graphicspath{./images/}
\DeclareUnicodeCharacter{2212}{-}
\title{Comparision Of Adversarial And Non-Adversarial LSTM Music Generative Models}

\renewcommand{\shorttitle}{Adversarial Music Generation}

\hypersetup{
pdftitle={Adversarial VS Non-Adversarial Music Generation},
pdfsubject={q-bio.NC, q-bio.QM},
pdfauthor={Moseli Mots'oehli, Anna Sergeevna Bosman, Johan Pieter De Villiers},
pdfkeywords={Music generation, MIDI, Generative adversarial networks, Long-Short Term Memory Neural Networks},
}

\author{ \href{https://orcid.org/0000-0002-9191-0565}{\includegraphics[scale=0.06]{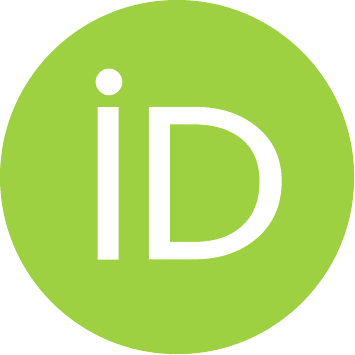}\hspace{1mm}Moseli Mots'oehli} \\
	Department of Information and Computer Science\\
	University of Hawai'i At Manoa\\
	Honolulu, HI 96822 \\
	\texttt{moselim@hawaii.edu} \\
	\And
	\href{https://orcid.org/0000-0003-3546-1467}{\includegraphics[scale=0.06]{orcid.pdf}\hspace{1mm}Anna Sergeevna Bosman} \\
	Department of Computer Science\\
	University Of Pretoria\\
	Pretoria, 0002 \\
	\texttt{annar@cs.up.ac.za} \\
	\And
	\href{https://orcid.org/0000-0003-2506-6594}{\includegraphics[scale=0.06]{orcid.pdf}\hspace{1mm}Johan Pieter De Villiers} \\
	Department of Electrical and Computer Engineering\\
	University Of Pretoria\\
	Pretoria, 0002 \\
	\texttt{pieter.devilliers@up.ac.za} \\
}

\begin{document}

\maketitle              % typeset the title of the contribution

\begin{abstract}
Algorithmic music composition is a way of composing musical pieces with minimal to no human intervention. While recurrent neural networks are traditionally applied to many sequence-to-sequence prediction tasks, including successful implementations of music composition, their standard supervised learning approach based on input-to-output mapping leads to a lack of note variety. These models can therefore be seen as potentially unsuitable for tasks such as music generation. Generative adversarial networks learn the generative distribution of data, and lead to varied samples. This work implements and compares adversarial and non-adversarial training of recurrent neural network music composers on MIDI data. The resulting music samples are evaluated by human listeners, their preferences recorded. The evaluation indicates that adversarial training produces more aesthetically pleasing music.
\end{abstract}

\section{Introduction}\label{sec:introduction}
Music composition, like most art forms, has for a long time been a skill specific to human beings. Music composition has an intuitive side to it necessary to determine which pitches create harmorny together, what chords can be played after a certain note, or what note progressions are in violation of intrinsic musical theory. With the recent successes in neural network modeling of predictive natural behaviour and generative models, there have been good applications of modelling note progression probabilities for music generation. The two dominant approaches to neural music generation are adversarial training \cite{Sutskever:Seq2Seq14,Liu:MissingMusic16,Yang:MidiNet17,Dong:MuseGan18}, and sequence-to-sequence recurrent networks \cite{Chung:EmpericalRNN14,Waite:LookBackRNN2016,Weel:RoboMozart17}, each with its merits. Although Wave-form representations have been shown to be a viable way to generate audio not necessarily specific to music \cite{Oord:WaveNet16}, it is symbolic representations that are favoured in literature for the task of music generation \cite{Mogren:CRNNGAN16,Yang:MidiNet17,Lerdahl:TonalGen83,Colombo:BachProp18,Chung:EmpericalRNN14}. Owing to the existing lack of out-right comparisons between adversarial and non-adversarial training for music generation, the aim of this study is to compare music samples generated by two generative models, one trained in an adversarial setting, and the other in a non adversarial setting, using musical instrument digital interface (MIDI) data.

This work strives to demonstrate two points, namely: (1) That generative adversarial networks (GANs) with long-short term memory neural network (LSTM) cells can be used to generate polyphonic music that is realistic, creative and pleasing to listen to, and (2) that generative adversarial models with LSTM cells perform better than an identical non-adversarial LSTM-based generator. An LSTM-based neural network is trained in an adversarial setting to generate music in MIDI format, and compared to an LSTM encoder-decoder network \cite{Cho:EnDe14}, that is not trained in an adversarial setting. Although adversarial training is much more complex in comparison to the encoder-decoder configuration for sequence-to-sequence models, GAN’s ability to model note progression by sampling a latent space leads to a more diverse generator. A Wasserstein generative adversarial network (WGAN) \cite{Arjovsky:WGAN17} is implemented instead of the maximum likelihood estimation (MLE) based GAN to ensure stable adversarial training. Although MIDInet \cite{Yang:MidiNet17}, a GAN based convolutional neural network (CNN) music generator, has been shown to produce better results compared to melodyRNN \cite{Waite:LookBackRNN2016}, which uses models that deploy recurrent neural network (RNN) cells, the two networks implement two different generator types, CNN and RNN, respectively, Thus the study provides no evidence to support the hypothesis that GAN training produces superior results in the music generation domain.
The musical data used in this work is in MIDI format. The simplicity inherent in pre-processing MIDI data as compared to pre-processing raw audio made MIDI a more suitable choice of music representation for the training data. To ensure music quality is not negatively affected by data representation, a common 2D state-matrix representation \cite{Shiebler:MusicRNNRBM17} of note progression is adopted for both training configurations. Although multi-track notes are captured, for the purpose of comparing the network’s ability to model note progression, it is trivial to also learn multi-track probability distributions, and instead assume a MIDI type 0 file explained in Section \ref{sec:Database} on decoding the resulting music and playback. The background of the architectures used in this study is also  explored, beginning with fully connected feed forward neural networks, activation functions, convolutions, recurrent networks, LSTM cells, and adversarial training. Standard music evaluation methods are used to come to a conclusion.

\section{Related Work}\label{sec:Related Work}
The goal of algorithmic music generation is to be able to develop systems that enable automation of the composition process, while still achieving results comparable to human generated music. Although music as an art form has existed for millennia, the earliest publications on algorithmic composition are only as old as 1960 \cite{Zaripov:AlgoMusic60}. It was only towards the mid 1970s that significant interest and research was put into algorithmic music generation. There are different approaches to algorithmic music generation such as using mathematical models (stochastic processes) \cite{Xenakis:FormalMusic92}, grammar based methods \cite{Lerdahl:TonalGen83}, learning algorithms \cite{Colombo:BachProp18,Dong:MuseGan18,Chung:EmpericalRNN14,Waite:LookBackRNN2016}, and evolutionary methods \cite{Alfonseca:Genetic07}. However, the most promising results have come from the learning algorithms in recent years, in particular deep learning neural networks. The focus of this study is on music composition using artificial neural network (ANN) learning algorithms \cite{Oord:WaveNet16,Yang:MidiNet17}.

A number of inventions in the deep learning domain contributed to majority of the work performed in neural music generation. The LSTM network \cite{Hochreiter:LSTM97} is well suited to successful learning of sequential data such as audio, and has the capability to recall notes generated a number of time steps back by solving the vanishing gradient dilemma that other RNNs suffered from. GANs \cite{Goodfellow:GANs14} are especially useful for generating realistic data while reducing over-fitting, and have been found to produce creative art \cite{Elgammal:CAN17,Juefei-Xu:GangOG17,YU:SeqGAN17}. Majority of the music generating neural networks to date are trained on either jazz or classical music, and only the piano track is used or all other tracks are played back on piano. SeqGAN \cite{YU:SeqGAN17} is a hybrid GAN between deep learning and reinforcement learning (RL), this model uses a RL generator agent to guide the generative learning. By using RL with the discriminator providing a reward function, seqGAN \cite{YU:SeqGAN17} is able to out-perform standard MLE based GANs in music generation. More recent work \cite{Hung:FreesoundGeneration21}, shows again that SeqGAN outperforms other methods on a  mucis generation bench-marking task based on the drum loops and FreeLoops datasets \cite{António:FreeLoops20}. Like authors of \cite{Hung:FreesoundGeneration21}, we implement a sequence based GAN, and consider subjective evaluation of the musical pieces produced.

 \cite{Colombo:BachProp18} proposed BachProp, an LSTM based network for learning note progression independent of note representation. They propose a three layered LSTM architecture to model notes, their timing and duration by conditioning the two other attributes on the current note per time step. Although BachProp uses a normalized MIDI representation of all training songs for training, they neglect to indicate how the network is representation invariant, as it assumes MIDI input data. Like in BachProp, \cite{Mogren:CRNNGAN16} introduced continuous-recurrent GAN (CRNNGAN) for the same task, and adopted a similar network structure with three stacked LSTM layers in the generator network to enable learning of high complexity notes, chords and melody with an additional MIDI feature (note intensity) over BachProp. Unlike BachProp \cite{Colombo:BachProp18}, CRNNGAN is trained in an adversarial setting. Due to their-three layered generator and continuous representation, CRNNGAN is limited to producing only up to three different tones per time step, hence produces music that is not rich in polyphony.

Authors of \cite{Chung:EmpericalRNN14} show that a gated recurrent unit neural network can be used to achieve results at least comparable to those of more sophisticated gated networks such as LSTM, keeping all training parameters equal. They train both models on piano-track MIDI data, and evaluate generated samples for polyphony. However, comparison over multiple datasets produced inconclusive results. MelodyRNN \cite{Waite:LookBackRNN2016} is a collection of RNN models (LookbackRNN and AttentionRNN) for polyphonic music generation trained on MIDI data. The LookBackRNN implements a look-back mechanism to help the network recall very long dependencies in generation, and the attentionRNN implements an attention mechanism \cite{Denil:Attend12} for increased note repetition to improve rhythm. WaveNet \cite{Oord:WaveNet16} introduced by the Google DeepMind team is a completely probabilistic network that uses the same architecture as PixelCNN \cite{Oord:PixelNet16} to generate raw audio waveforms from a dataset of mp3 files with multiple tagged genres. WaveNet was developed mainly for the task of text to speech synthesis, and uses multiple layers of time dilated convolutional networks instead of more traditional and suiting sequence models such as RNNs. The DeepMind team do this to avoid the long training time required for RNNs. However, WaveNet’s generator is not trained in an adversarial setting, and no quantitative results on music generation are reported by the authors.  Residual-CNNs are successfully implemented in the GAN training framework in \cite{Chen:AutomaticGANDJ22}, where the task if to generate between track transitions similar to those made by a disco jockey (DJ). The authors of \cite{Chen:AutomaticGANDJ22} also define a custom subjective evaluation methodology to conclude the musical pieces produced are competitive with those of human DJs. MidiNet \cite{Yang:MidiNet17} expands Wavenet this work using MIDI files instead, and train CNNs in an adversarial setting. Both networks learn note progression by sequentially conditioning future notes on the distribution of previously generated notes by using dilated convolutions. Although MidiNet outperforms standard RNNs, and produces music that is considered more varied and pleasing to listen to than both the LookbackRNN and AttentionRNN, it is unclear as to how it would compare to LSTM and gated recurrent unit (GRU) based networks which are superior to standard RNNs on very long sequence tasks. This work implements an encoder-decoder LSTM network in the same fashion as BachProp, and an LSTM based GAN as in \cite{YU:SeqGAN17,Mogren:CRNNGAN16,Colombo:BachProp18}.  \cite{Edirisooriya:OMR21} also use the encoder-decoder architecture for evaluation of end-to-end polyphonic optical music recognition models.

The majority of the work mentioned above use MIDI training data, and only a few train from raw waveforms. The most common way of representing the MIDI features to be learned is using a 2D matrix \cite{Dong:MuseGan18} of binary entries where pitch is on one axis, and the other axis represents time in ticks. In some cases the real continuous values from the MIDI messages are entries in the 2D matrix \cite{Mogren:CRNNGAN16,Weel:RoboMozart17}. Some work has been done in learning multi-track note progressions, although results show lack of synchrony and cross track dependency learning. \cite{Chu:PiMusic17} implement a sequence-to-sequence(Seq2Seq) model of four stacked LSTM layers to model multi-track note progression with each LSTM cell generating its own track’s output. Other notable neural network approaches for music generation include Variational Autoencoders \cite{Alexey:RVAEHistory20,Roberts:HVAEMusic17}, restricted Boltzmann machines \cite{Vincent:TemporalDependencies12}, and deep belief neural networks \cite{Bretan:UnitSelection17} to learn note progression probabilities that boost rhythmic scores. They show how the resulting model can be used as a prior distribution for training an RNN for polyphonic music.

In \cite{Shuqi:DaiControllable21}, the authors introduce ``MusicFrameWorks", a hierarchical music representation structure that allows for the treatment of melody, rhythm and chord progression as separate but conditional tasks. While, unlike all the work listed thus far, MusiFrameWorks implements two transformer-based networks \cite{Vaswani:AttentionIsAll17} that have been shown to outperform LSTMs in the natural language processing (NLP) realm, their musical representation falls short when it comes to capturing polyphony and harmonic progression. As in most of the listed literature, MusicFrameWorks relies on subjective evaluation methods.

\section{Database}\label{sec:Database}

\begin{table}
 \begin{center}
 \begin{tabular}{|l|l|l|l|l|}
  \hline
  Channel & Note & Time & Type & Velocity \\
  \hline
  \hline
  0 & 39 & 0 & note\_on & 80 \\
  0 & 58 & 0 & note\_on & 80 \\
  0 & 46 & 0 & note\_on & 80 \\
  0 & 61 & 0 & note\_on & 80 \\
  0 & 70 & 0 & note\_on & 80 \\
  0 & 70 & 240 & note\_off & 64 \\
  9 & 39 & 0 & note\_on & 80 \\
  9 & 39 & 480 & note\_off & 64 \\
  1 & 46 & 0 & note\_on & 80 \\ \hline
 \end{tabular}
\end{center}
 \caption{A sample MIDI file with note action messages presented in tabular form. The “Note” column represents pitch. Note type communicates which notes are on at each point in time, and on which of the 16 channels the message is transmitted}
 \label{tab:MIDI_EXAMPLE}
\end{table}

Musical instrument digital interface (MIDI) is a set of standards that outline how to connect digital music instruments, so they can communicate using a messaging layout that is standardized. These messages are termed MIDI messages, and encode instructions that can be decoded to produce sound by any digital instrument that conforms to the MIDI standard. There are several MIDI message types: “note on” messages, “note off” messages, meta messages, control change messages, program change messages, and tempo change messages.To assemble a fully functional MIDI file, all these messages are required, but for the purpose of modelling note progression and suitability to evaluate the model’s ability to generate audio, “note on” and “note off” messages on their own are sufficient, with other messages being set to default values. Each note event message contains several attributes that accompany it. Table \ref{tab:MIDI_EXAMPLE} shows how note action messages are stored in a MIDI file. Since MIDI files are compact relative to the raw audio represented, there are numerous training dataset sources available online to choose from for the task of neural music generation. The Lahk MIDI dataset \cite{Raffel:LearningBasedMethods18} is one of the most commonly used for this purpose. The dataset is a collection of 176,581 unique MIDI files of different genres and composers. Of the seven versions of the dataset, the “clean MIDI subset”version containing 17,257 song with filenames indicating song titles and artist was used. This dataset is of size 224MB compressed and 770MB uncompressed. Due to memory constraints, only 289 distinct polyphonic songs from 10 composers were used for training, comprising only 10MB of disk space. Table \ref{tab:dataset} gives a statistical summary of the training data used for the different composers.

\begin{table}
 \begin{center}
 \begin{tabular}{|l|l|l|}
  \hline
  Composer & Number of Songs & Number of Notes \\
  \hline
  \hline
  Beethoven & 29 & 158978 \\
  Billy Joel & 117 & 947296 \\
  Borodin & 7 & 24260 \\
  Diana Ross & 3 & 39022 \\
  Elton John & 31 & 330044 \\
  Elvis & 5 & 29776 \\
  Frank Sinatra & 45 & 268180 \\
  Liszt & 16 & 53366 \\
  Mendelssohn & 15 & 79516 \\
  Mozart & 21 & 81470 \\
  \hline
  Total & 289 & 2011908\\ \hline
 \end{tabular}
\end{center}
 \caption{Training dataset description by composer and number of songs. The number of Notes represent both note on and off messages to be learned}
 \label{tab:dataset}
\end{table}

\section{Methods}\label{sec:Methods}
The note progression state-matrix representation algorithm used in this work is adopted from \cite{Shiebler:MusicRNNRBM17}. It takes a standard MIDI file as input and transforms the “note on” and “note off” messages into a matrix of binary entries. In this representation, only information pertaining to note messages is kept, that is, whether a note is on or off, the timing and duration of the note. The algorithm consists of two separate processes, one for encoding a MIDI file into the note state-matrix representation, the other for decoding and transforming an existing note state-matrix into a valid MIDI file that can be transcribed by any MIDI enabled device. The encoding and decoding processes are discussed in the following section.

\subsection{Encoding}\label{subsec:Encoding}
Starting with a MIDI file with textual and numerical data, the aim is to extract and transform into binary all note information to be used as training data for music generation. For each file, all messages relating to the same pitch are first lined up in ascending order of their delta times, with each present pitch having a list that starts at tick time 0. Adding up all the delta times on the pitch that plays last in the song produces the duration of the song. From the assembly of messages per pitch present in a song, three important attributes, namely pitch, time and note message type, are extracted and used to create the state-matrix representation of the song.

\begin{figure}[!htb]
\caption{ Visualization of sample MIDI files in the note state-matrix representation.}
\centering
\begin{minipage}[b]{0.47\textwidth}
    \includegraphics[width=\textwidth]{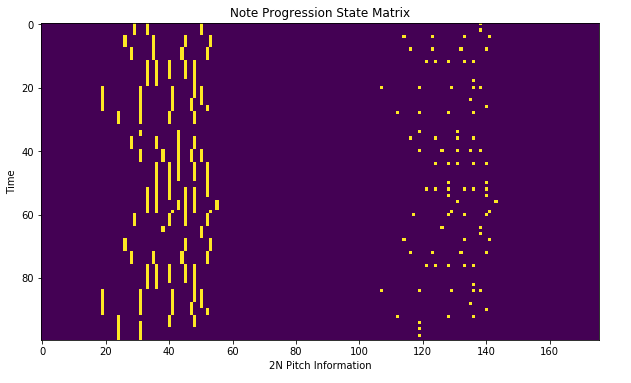}
\end{minipage}
\hfill
\begin{minipage}[b]{0.47\textwidth}
    \includegraphics[width=\textwidth]{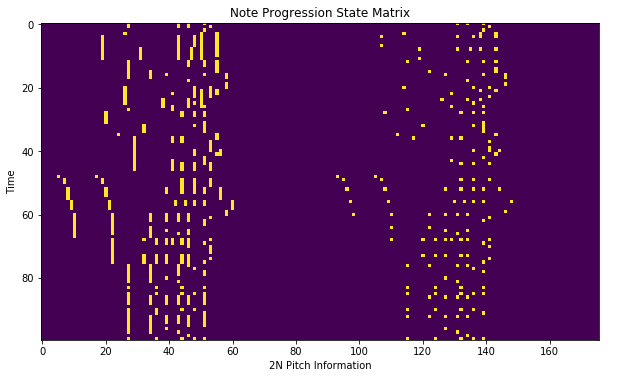}
\end{minipage}
\label{fig:MIDI_VISUALIZATION}
\end{figure}

Extracting the pitch and note message type are fairly straightforward operations since they are explicitly contained in each of the ordered messages. The $M \times 2N$ matrix is constructed so that the $M$ rows represent tick times. Of the $2N$ columns representing the pitch information for $N$ possible pitch values, the first $N$ columns uniquely identify each pitch value, and the next $N$ indicate whether the pitch was played in the previous time step or not. $N$ represents the number of possible pitches which is 128 by default. The entries into the matrix are binary to indicate the state of the note at each tick time on all pitches. Figure \ref{fig:MIDI_VISUALIZATION} shows the binary heat-maps of the state-matrix representations of notes in 2 songs. This representation allows for easy expression of multiple chords, hence allowing the models to capture as much information on polyphonic note progressions as possible. One complication with the state-matrix representation is that it creates an imbalance in the prediction space. This is because each time-step in the matrix contains a very small proportion of note-on signals as compared to note-off signals for the 128 available pitches. Table \ref{tab:Data_Imbalance} shows the proportion of prediction instances representing positive note activations and negative activations for the training dataset used in this study.

\begin{table}
 \begin{center}
 \begin{tabular}{|l|l|l|}
  \hline
  Note Message Type & \# Messages & Percentage \\
  \hline
  \hline
  Note-off & 5924916 & 96.63\% \\
  Note-on & 199884 & 3.37\% \\
  \hline
  \bf Total & \bf 6124800 & \bf 100\% \\ \hline
 \end{tabular}
\end{center}
 \caption{Number of prediction instances/note messages in the training dataset showing the data class imbalance inherent in the state-matrix representation used in this work.}
 \label{tab:Data_Imbalance}
\end{table}

Owing to the imbalance shown in Table \ref{tab:Data_Imbalance}, a model that simply predicts note-off messages for all 128 pitch values at each time step will achieve an accuracy score of approximately 96\%. BAcc, which is calculated as the weighted prediction accuracy between the number of prediction classes, is not prone to the same problem as prediction accuracy \cite{Henning:BalancedAccuracy10,Vicente:Indexbalanced09,Chongsheng:ImbalancedFeature22}, and so it is used for both training and testing in this work. To create the training dataset, all training and test MIDI files are passed through this process to create a collection of note progression state matrices, which results in a three dimensional matrix. This is ideal since the models explained in Section \ref{sec:Models} expect a three dimensional input.

\subsection{Decoding}\label{subsec:Decoding}
The decoding of a note state-matrix is the reverse process of the encoding process, and as such is highly dependent on the encoding. This process accepts a binary state-matrix of dimension defined in the encoding phase, as input, and transforms the matrix into a valid MIDI file. Without this process, it would be hard to evaluate the quality of audio samples generated by both the encoder-decoder and WGAN models discussed in Section \ref{sec:Models}. To be able to generate a valid transcribable MIDI song, each observation in the state-matrix is written as a note message in the MIDI file comprising the following attributes as a minimum requirement: channel, pitch, time, and velocity. Note message type is also a necessary attribute for valid MIDI transcription. For play-back purpose, a meta tempo message is also sent to the file before all other note messages. However, the models compared in this work only model pitch progression states, and so the velocity, channel and playback tempo are kept constant at 70, 1 (Acoustic Grand Piano) and 120 respectively to ensure the audio is loud enough and of standard pace. Delta time of each message is calculated based on the number of ticks between the current message and the previous message of the same pitch, scaled on the constant file tempo. In the case that an observation in the state-matrix representation contains information about more than one pitch, multiple MIDI messages are generated from this one observation with the same delta time. The time attribute $d_{t,s}$ for pitch $s$ is extracted only after all note progression states in the matrix are determined using formula \ref{eq:dst} below:

\begin{equation}\label{eq:dst}
    d_{t,s} = \frac{v_{s_{t}}-v_{s_{t-1}}}{\tau}
\end{equation}

\noindent
where $v_{s_{t}}$ is the tick time of the current message containing instruction for state $s$, $v_{s_{t-1}}$ is the tick time of the previous message containing information on the same state, and $\tau = 120$ represents fixed standardizing tempo. Decoding pitch is more complex than all the other required attributes. In the note state-matrix representation of binary entries, the first $N − 1$ columns represent pitch activations, and the next $N$ to $2N − 1$ represent pitch retention of each observation. A value of 1 in the first $N−1$ columns is recorded as a “note on” message, and a value of 0 denotes a “note off” message for the pitch encoded by the column. In the next $N$ to $2N − 1$ columns, a value of 1 instructs the MIDI transcriber to activate the corresponding pitch in column in the first $N − 1$ columns, while a value of 0 releases the pitch. Activations of a pitch of constant velocity for continuous time steps until release constitute an elongated pitch press to a human listener.

When the state-matrix attributes have been extracted and written to a MIDI file, The decoding process assumes a MIDI file type 1, where all messages are written to one track.

\section{Models}\label{sec:Models}
The encoder-decoder and the adversarial generative models are presented in  Sections \ref{subsec:Encoder-Decoder LSTM} and \ref{subsec:LSTM WGAN}, respectively. Their architectures are explored together with some of the key structural training parameters such as activation functions, number of stacked LSTM layers, and optimizers used. Since the training configurations are different, the last part in each configuration explains how the preprocessed note state-matrix data is presented to the network for training.

\subsection{Encoder-Decoder LSTM}\label{subsec:Encoder-Decoder LSTM}
The encoder-decoder configuration \cite{Cho:EnDe14}, is well suited to the modelling of sequential tasks such as text translation, textual question answering, text summarization and music generation. In this configuration, two networks joint end to end with only one objective function are trained so that the first network (encoder) is fed all the training sequences and encodes what it has learned in a low dimensional vector. The decoder then learns the mapping from this encoded “thought vector” to the desired sequence of outputs. The decoder does this by optimizing a loss function, and is trained using Back propagation through time (BPTT) \cite{Werbos:BPTT90}. Since the two networks are connected, gradients from the decoder are propagated all the way back to the encoder, so the encoder also improves its encoding process.

\begin{figure}[!htb]
\caption{The encoder bidirectional LSTM network. Inputs $S_{t+i}$ represent note pitch information per time-series observation (tick) pulled from the 2D note progression state-matrix.}
\label{fig:ENCODER_LSTM}
\centering
\includegraphics[width=0.8\textwidth]{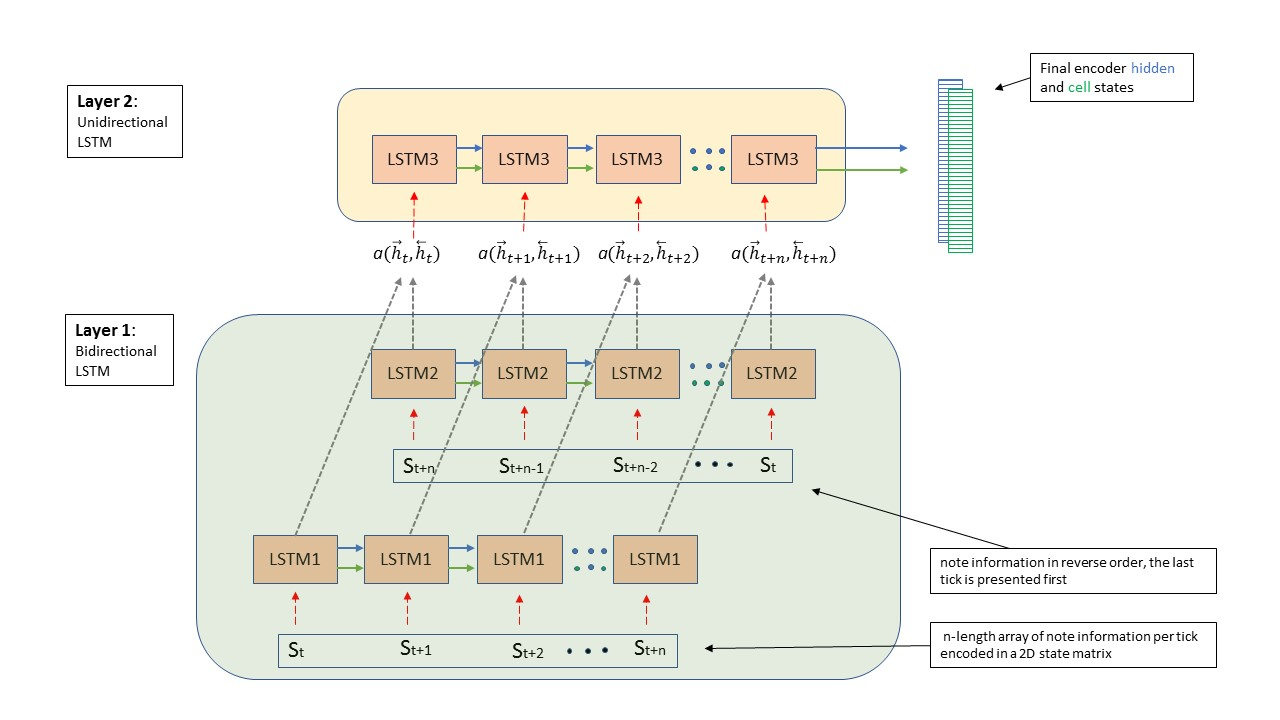}
\end{figure}

It has been shown that stacking LSTM cells results in improved performance \cite{Mogren:CRNNGAN16,Cui:StackedLSTM17}, and so three stacked LSTM cells in both the encoder and decoder networks were implemented. The reason for stacking LSTM layers is to capture multiple levels of abstraction that could be inherent in the temporal data. For music generation, the sequence of note information encodes not just note progressions per track, but also chord progressions and phrases that contribute to the rhythm and melody of the input and output. Capturing both the forward and backward flow of input during encoder training has shown to significantly improve the cell state’s memory retention and subsequently the quality of generated sequences \cite{Schuster:BiLSTM97,Cui:StackedLSTM17} by the decoder. For this reason, one of the encoder layers reads the input sequence backwards, the other forward, and the third takes the aggregation function $a(h_{1}, h_{2})$ result of the other two layers’s hidden states as input to finally produce the final hidden state $V_{T} = \theta_{3}(a(\vec{h},\vec{h}))$, where $\theta_{i}(.)$ comprises all LSTM equations. Figure \ref{fig:ENCODER_LSTM} shows bidirectional stacked encoder network. This bidirectional input approach was successfully implemented for music generation in \cite{Mogren:CRNNGAN16,Liu:MissingMusic16}.

The first two LSTM layers can be considered to be on the same hierarchical level, hence making up just a single bidirectional layer. The third LSTM is stacked over the bidirectional layer. The Aggregation function can either be a summation, multiplication or concatenation operation over the hidden states of the forward and backward reading LSTM cells. Each LSTM cell internally consists of standard LSTM operations.The decoder comprises 3 stacked unidirectional LSTM cells with 256 neurons each, and sigmoid output activation. Unlike the encoder, the decoder is used in two modes, notably: training and inference. These two modes differ by the manner in which information flows from input to output until termination of decoding. During training, the decoder hidden and cell states are initialized using the encoder’s final states. Since an LSTM cell expects three inputs, a zero vector $st_{0} = \vec{0}$ is presented as a primer to the decoder, and the first predicted note progression state $\hat{st}_{1}$ is generated by the network. During training, the real musical note progression states $st_{t}, st_{t+1}, ..., st_{t+n−1}$ are presented as input per time step to predict the orderly sequence of note progression states $\hat{st}_{t+1}, \hat{st}_{t+2}, ..., \hat{st}_{t+n}$.

\begin{figure}[!htb]
\caption{The decoder unidirectional LSTM network. Inputs St+i represent note pitch information per time-series observation (tick) pulled from the 2D note progression state-matrix.}
\label{fig:DECODER_LSTM}
\centering
\includegraphics[width=0.9\textwidth]{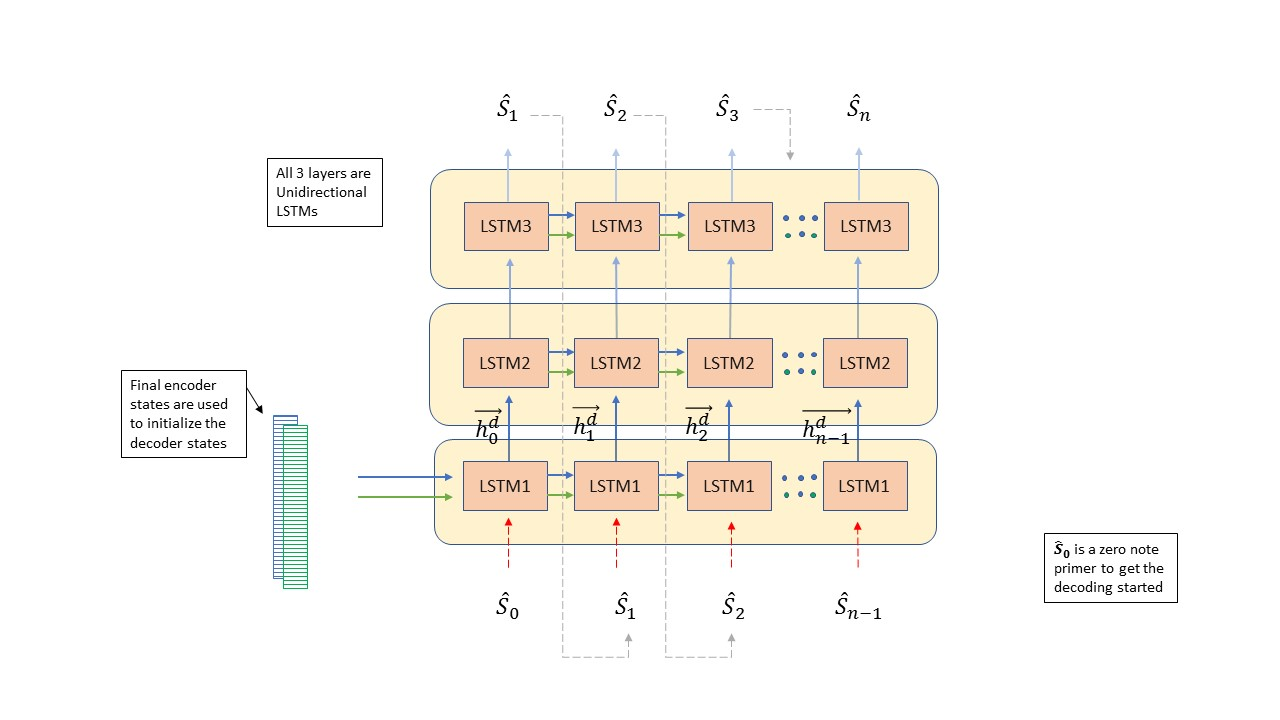}
\end{figure}

As in all other optimization-based networks, the predictions $\hat{st}_{t+i}$ are compared to the actual note states $st_{t+i}$ to calculate the loss. The RMSprop \cite{Mukkamala:RMSProp17} optimizer is used to minimize the overall encoder-decoder network cross entropy objective. RMSprop is the recommended choice for training RNNs according to Keras as it speeds up training, and has been successsfully used in training RNNs on temporal data \cite{Lopyrev:15,Kim:KimDeepJaz17,Roberts:HeiracicalLVM18}. Post training, a short sequence of notes is presented to the encoder which passes on a low dimensional vector representation of its final hidden state to the decoder. Inference proceeds with a primer input, first note state prediction is passed as input into the decoder in the next time-step, and the process continues recursively until the desired number of time-steps of output is reached as depicted in Figure \ref{fig:DECODER_LSTM}. In all the LSTM cells, dropout(0.3) is used for regularization. Since each song is represented by a $T \times 2N$ matrix, where $T$ represents number of ticks/time-steps and $N$ the number of allowable pitches to model, all LSTM cells have layers containing $2N$ hidden units. The algorithm used for extracting note information from the MIDI file and creating the note state-matrix representation was discussed in Section \ref{subsec:Encoding}. Encoder-decoder networks are sequence-to-sequence models, and so accept a sequence of priming notes in order to be able to produce an output sequence. In music generation, as in any other sequence-to-sequence task, the more data presented to the encoder network, the better the quality of information available to the decoder network. However, longer sequences can have negative effects on learning due to vanishing gradients. For encoder-decoder music generation, all songs are limited to $N$ ticks; $m$ of these, where $m < N$, are presented to the encoder and $n$, with $m < n < N$, to the decoder during training. Note that $m + n = N$. All training songs are split in this manner and presented to the network in batches of eight. During inference, the network expects a primer sequence of $m$ notes and generates $n$ note progression states probabilities, and a threshold value of $0.5$ is used to turn notes on. The threshold is set to $0.5$ because the values being predicted are probabilities of a note being played. Once there is a sequence of notes generated, it is presented to the MIDI note state-matrix decoding method in Section \ref{subsec:Decoding} to produce a MIDI file that can play on any MIDI enabled device. In the following section, the WGAN implementation is discussed.

\subsection{LSTM WGAN}\label{subsec:LSTM WGAN}
Wasserstein distance or earth mover (EM) distance is a measure of the amount of work required to move one probability distribution to another. While traditional GAN seeks a density distribution $P_{\theta}$ that maximizes the likelihood of samples from the distribution $P_{r}$ to be modelled, WGAN minimizes the Kullback Leibler (KL) divergence distance which is a reasonable approximation to EM distance \cite{Arjovsky:WGAN17}. EM distance can be approximated by the equation: 

\begin{equation}
W(P_{r},P_{\theta}) = inf_{\gamma \in \prod (P_{r},P_{\theta})} E_{(x,y)\sim \gamma}[\|x-y\|]
\end{equation}

\noindent
Where $P_{r}$ and $P_{\theta}$ represent the unknown target distribution and the learned parameterized density distributions respectively. $X, Y  \in \mathbb{R}_{d}$.
This objective has properties that ensure convergence in situations where other distance measures fail to converge. An example is the case in traditional GANs where the support of $P_{\theta}$ is drawn from a low dimensional latent space. If the overlap between the latent space and  the real data generating distribution $P_{r}$ is significant enough, most distance measures are invalid or deem the distance infinite. As a result of this, WGAN is much more stable compared to GAN, and needs less architectural hyper-parameter tuning of the generator and discriminator. As such, WGAN is an improvement over standard GAN training. The WGAN implementation in this work consists of an LSTM generator and discriminator. Since the goal is to compare adversarial training to encoder-decoder configured training, both the generator and discriminator network architectures are identical to the decoder and encoder respectively in terms of the number of LSTM layers, number of neurons in each layer, and activation functions except for the output activation. Other training hyper-parameters such as learning rate and stopping criteria are left to vary per training configuration, and is discussed in Section \ref{sec:EXPERIMENTS}. Below is a detailed description of the WGAN implementation used in this study.
The generator $G(z)$ in WGAN is similar to any other generative model that generates melodies from random noise. The specific implementation in this study has three stacked unidirectional LSTM layers with 256 neurons in each layer, an input layer, a fully connected layer before the output, and an output layer. The input layer accepts a matrix of latent variables of size 256 for each generation time-step from the standard normal distribution. The first LSTM layer takes this latent matrix as input per time-step and passes the information on to the higher layers, and the state output per time-step is passed on to the next time-step to generate a sequence of outputs. The outputs should transform into realistic note state-matrix probabilities over the course of training. The output of the generator network is forced to have the same dimensions as those of the state-matrix representation of the real data.

\begin{figure}[!htb]
\caption{WGAN architecture for image generation. Source: \cite{Hui:WGANGP18}}
\label{fig:WGAN_ARCHITECHTURE}
\centering
\includegraphics[width=0.8\textwidth]{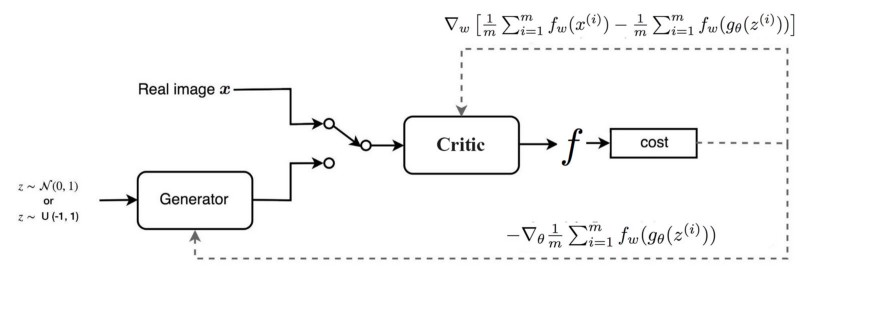}
\end{figure}

In traditional GAN training, the discriminator $D(x)$ is an adversary to $G(z)$ as they are trained in competition. The $D(G(z))$ in WGAN acts more in partnership with $G(z)$ in that, $D(G(z))$ is trained to optimality much faster than $G(z)$, and so $D(G(z))$ is able to provide important loss information to $G(z)$ very early in training to speed up $G(z)$’s convergence. In this manner the discriminator is comparable to the encoder-decoder network in that it provides the necessary information for the generator’s optimal training. The $D(G(z))$ implementation has the same number of LSTM layers, including one bidirectional layer as the encoder. The network expects input with dimensions consistent to the note state-matrix representation of the MIDI files, and produces a scalar output to ensure the distance between real and fake sample outputs is as large as possible. This is unlike traditional GANs that produce a probability that the song is a sample from the real data generating distribution. For this purpose a linear output activation function is used for $D(G(z))$. The scalar output helps in calculating the EM distance, which is used in $D(G(z))$'s and $G(z)$'s loss functions. The discriminator takes both real note samples and random noise as inputs together with their labels; 1 for real and -1 for fake samples. Random noise is passed through a partially trained generator at that point in time for each epoch, and the resulting samples represent generated music. To ensure $D(G(z))$ is always more informed than $G(z)$ to be able to guide $G(z)$’s loss to optimality, $D(G(z))$ is trained five times more for every single epoch of $G(z)$. Figure \ref{fig:WGAN_ARCHITECHTURE} shows the structural setup of the WGAN configuration with $D(G(z))$ and $G(z)$ as explained above applied on image generation.

\section{Evaluation Methods}\label{sec:Evaluation Methods}
The training and validation errors of the models were recorded and are be reported in Section \ref{sec:Resuts}, however they do not represent a complete measure for the performance of a generative algorithm. In generative models, the goal is to be able to learn the generative distribution underlying the samples used during training, while at the same time not over-fitting on these few examples. This means that it is possible to have a model that has average training and validation accuracy produce much more realistic and creative musical samples than one with a very high training accuracy, as music is an art form and is very subjective. Majority of existing literature in neural music generation relies on subjective evaluation methods, with human listener surveys being the most common approach \cite{Oord:WaveNet16,Mogren:CRNNGAN16,Colombo:BachProp18,Lackner:MelodyLSTM16}. For this study, a survey was conducted where listener impression scores from 10 individual volunteers were collected. The scores give a subjective opinion of the listeners impression of the samples generated by both the encoder-decoder and the WGAN generators.

The 10 human evaluators of the samples were chosen at random from a group of friends and each was contacted through Whatsapp, messaging to be asked to participate in the study. This method of contact was especially suitable, as it allowed for quick and easy access to individuals that would likely agree to participate in the study. The fact that the social messaging application has functionality for sharing audio was also one influential factor for the choice of survey design. All the volunteers were aged between 20 and 40 with an uneven split of gender (6 male, 4 female). They were notified that they will receive eight distinct pairs of audio samples not longer than 2 minutes in length that are both generated by a learning algorithm. They were then asked to rate each of the samples by giving it a score in the range [0, 5], (where 0 is completely random noise and 5 is a good song), to express how much they enjoyed listening to the song and how rhythmic and melodic the music samples are. A numerical score was collected for each sample in the comparison pair. This enabled calculation as a percentage of the number of times samples from one model are preferred over those from the competing model. This together with the median represent a good measure of centrality of the scores to ensure the analysis is not influenced by outliers. Each participant received a group of eight randomized pairings of encoder-decoder and WGAN samples to compare and score, resulting in a total of 80 ratings. Microsoft Excel was used to randomize the samples and to assign them to the volunteers. The importance of mentioning to the volunteers that both samples were algorithmically generated was to ensure they do not score the samples by comparing them to what they define as really good music generated by expert human artists. The volunteers were not informed as to which samples were generated by which model. Some of the volunteers, although not asked, were able to provide textual explanation for their preference in the samples, which provided for a better comparison of the two generative models. Once the scores were collected, the mean opinion score (MOS) \cite{Huang:DL4Music16} test was conducted to get to the conclusion that answers the research questions. For each WGAN sample $S_{i}^{w}$ and encoder-decoder sample $S_{i}^{ed}$, MOS $q(S)$ over all 10 volunteer ratings $r_{k}(S)$ is defined as:

\begin{equation}\label{eq:qsi}
    q(S_{i}) = \frac{1}{10}\sum_{k}^{10}r_{k}(S_{i})
\end{equation}

\noindent
where the rating $r_{k}(s_{i})$ is the rating given to sample $S_{i}$ by volunteer number $k$ of 10 volunteers.The mean generator quality $Q(S_{g})$ of a group of samples from the same generator g, is defined as the average MOS given by :

\begin{equation}\label{eq:Qsg}
    Q(S_{g}) = \frac{1}{8}\sum_{i}^{8}q(S_{i}^{g})
\end{equation}

\noindent
where $S_{i}^{g}$ for $i = 1, 2, 3, ..., 8$ represents samples from generator $g$. Since there is subjectivity in the measure of quality, it is important to quantify how much variation there is in the recorded sample and model qualities using the standard deviation. This also expresses how much confidence is placed in the estimate of quality used, where a high standard deviation represents low confidence in the accuracy of the estimate, and a low deviation represents high confidence. The two standard deviation estimates for MOS and mean generator quality are expressed below respectively:

\begin{equation}\label{eq:sigma_q}
    \sigma_{q(S_{i})} = \sqrt{\frac{\sum_{k}^{10}(r_{k}(S_{i}) - q(S_{i}))^2}{10-1}}
\end{equation}

\begin{equation}\label{eq:sigma_Q_big}
    \sigma_{Q(S_{g})} = \sqrt{\frac{\sum_{i}^{8}(q(S_{i}^{g}) - Q(S_{g}))^2}{10-1}}
\end{equation}

\noindent
with $q(S_{i})$ and $Q(S_{g})$ given by Equations \ref{eq:qsi} and \ref{eq:Qsg}, respectively. Although one generator may have a higher MOS than the other, tests have to be performed to ensure the MOS estimate of quality is not negatively influenced by outliers, and that it indeed has a different and higher median opinion score. The Wilcoxon Signed-Rank t-test \cite{Wilcoxon:PiMusic17} was used to perform the test of equal medians in ranked pair data, discussed in the next section.

\subsection{Wilcoxon Signed-Rank T-Test}\label{subsec: Wilcoxon Signed-Rank T-Test}
The Wilcoxon signed-rank t-test is a statistical hypothesis test for ranked and paired data that assumes no predefined population distribution over which the data is sampled from. Wilcoxon signed-rank t-test is used for comparing related pair samples under the hypothesis that the median difference between the samples is zero. The test makes the following assumptions about the data:

\begin{itemize}
  \item The data observations are paired samples from the same population.
  \item Each pair is chosen randomly and independently.
  \item The observations are measured on an ordinal, not necessarily nominal scale.
\end{itemize}

The assumptions above are met by the opinion score data collection process to a suitable extent in that: 

\begin{enumerate}
    \item The ranked samples come from the same population of ranking volunteers.
    \item The pairs were chosen randomly, though independence in this case is subjective as it is important to ensure all samples from both models were evaluated by the same number of volunteers. This was achieved by random selection without replacement.
    \item  Although the scores indicate by how much one sample is better than the other, hence violating the ordinality assumption, a transformation is applied to the scores to ensure only the ordinal aspect of the scores is used.In this transformation, the score pairs are compared, and a new indicator feature is constructed that assumes the following values: positive $(+)$ if the second sample has a higher score, zero if the scores are tied, and negative $(-)$ otherwise.
\end{enumerate}

\noindent
With this new sign transformed data, all the test assumptions are satisfied. The null and alternative hypothesis are given by:\\
\indent $H_{0}$: The median score difference between the paired samples is zero.\\
\indent $H_{1}$: The median score difference is not zero.

\vspace{5mm} %5mm vertical space
\noindent
Let $ed$ and $wg$ represent the encoder-decoder and WGAN models, respectively, and $sgn$ represent the sign function. With data observations $sgn(r_{ed,i} − r_{wg,i})$, all tied pairs are excluded from the initial sample of size N to a reduced test sample of size $N_{r}$. The original sample pairs are ordered in ascension according to the absolute differences $r_{ed,i} − r_{wg,i}$ of the original captured scores with the smallest difference ranking first as 1 and all ties receiving the average rank of the positions they span. With these new pair ranks $R_{i}$, the Wilcoxon signed-rank t-test statistic is calculated as follows:

\begin{equation}\label{eq:wilcoxon_statistic}
    W = \sum_{i}^{N_{r}}(sgn(r_{ed,i} − r_{wg,i})\cdot R_{i})
\end{equation}

Note that using the reduced sample size $N_{r}$ is equivalent to using the original sample size $N$, since all tied pairs result in a zero sign, hence not contributing to $W$. For $N_{r} \geq 10, W$ is asymptotically normally distributed, thus the z-score can be calculated as follows:

\begin{equation}\label{eq:z_score}
    Z = \frac{W-0.5}{\sigma_{W}}
\end{equation}

\noindent
with:

\begin{equation}\label{eq:sigmaW}
    \sigma_{W} = \frac{N_{r}(N_{r}+1)(2N_{r}+1)}{6}
\end{equation}

The null hypothesis $H_{0}$: equal medians, is rejected in favour of $H_{1}$: unequal medians, if $z > z_{critical}$. Rejecting the null hypothesis would mean the two models have generated sample scores with statistically different medians, and so there is more confidence that the contribution of outliers in the mean model and opinion scores is trivial. Based on the results of the MOS, mean generator quality, and the Wilcoxon signed rank test, the generator with the higher mean generator quality is considered to produce music that is more aesthetically pleasing to listen to, given the null hypothesis of equal medians is rejected. The volatility estimates given by Equation \ref{eq:sigma_Q_big} are used as a measure of the confidence in making the conclusion based on the MOS estimates above. The textual comments collected on some samples are analysed to get more insight into how people perceived the music. However, due to the lack of correct collection of these comments, no computational sentiment analysis is performed data, but a human opinion sentiment analysis of the text is provided.

\section{Experiments}\label{sec:EXPERIMENTS}
The implementation configurations of the MIDI state-matrix representation are briefly discussed, followed by a discussion on the training parameters for both generative architectures. Finally, we focus on the generation of music samples used in evaluating the encoder-decoder and WGAN models. All computation relating to the data and models was performed on a 7th generation core I7 intel processor, two Gigabytes Nvidia GeFORCE GPU personal computer with 16 Gigabytes of RAM.

\subsection{MIDI Representation}\label{subsec:MIDI Representation}
\noindent
Of the 128 possible pitch values, existing implementations use only 88 pitch values between 21 and 109 since all other pitches outside this range are inaudible to the human ear. This in effect reduces the state-matrix dimensionality, and overall model complexity. In this work, all 128 pitch values are used to avoid the added pre, and post processing in using a reduced note representation. For the models trained below, changing from 88 pitch values to 128 pitch values had an increase in model parameters of only 16\% and 22\% for the WGAN and the encoder-decoder models, respectively

\subsection{Model Hyper-Parameters}\label{subsec:Model Hyper-Parameters}
Below is the final list of models training hyper-parameters used in the study. A brief explanation of how they affected learning then follows per model.

\subsubsection{Encoder-Decoder}\label{subsubsec:Encoder-Decoder_Experiments}
\noindent
The training hyper-parameters for the encoder-decoder LSTM neural network are shown in Table \ref{tab:encoder-decoder_hyper}.

\begin{table}
 \begin{center}
 \begin{tabular}{|l|l|}
  \hline
  \bf Hyper-parameter & \bf Final Value \\
  \hline
  \hline
  Learning Rate & 0.001 \\
  Optimizer & RMSprop  \\
  Dropout Probability & 0.3  \\
  Training Epochs & 300  \\
  Batch Size & 32  \\
  Hidden Layer Size & 256 \\
  Gradient Clipping Max & 2.0 \\
  Train test Split & 80:20 \\
  \hline
 \end{tabular}
\end{center}
 \caption{Final training hyper-parameters for the encoder-decoder LSTM.}
 \label{tab:encoder-decoder_hyper}
\end{table}

\noindent
The learning rate was set low to ensure smoother tracking down the loss function as higher learning rates converge quicker, but the quality of music generated was bad hinting at convergence to a local minimum. Dropout \cite{Srivastava:Dropout14} as a regularization technique added more training stability and reduced over-fitting. Other parameters such as gradient clipping and batch size were determined using five-fold cross validation with all other parameters fixed. Table \ref{tab:Mean_5_fold_CV} shows mean BAcc scores over the five-fold cross validation (CV) for different learning rates and batch sizes. The model’s training mini-batch size and learning rate were  from the 5 fold CV, and they are the values that achieved the highest BAcc score.

\begin{table}
 \begin{center}
 \begin{tabular}{|l|l|l|l|}
  \hline
  &\multicolumn{3}{|c|}{\bf Batch Size} \\
  \hline
  \bf Learning Rate & 5 & 8 & 10\\\hline \hline
  0.0005 & 71.2\% & 56.6\% & 51.1\% \\
  0.001 & 74.1\% & 69.6\% & 60.3\%  \\
  0.005 & 73.6\% & 73.2\% & 72.8\%  \\\hline
 \end{tabular}
\end{center}
 \caption{Mean 5-fold CV balanced accuracy scores for different  batch sizes and learning rates.}
 \label{tab:Mean_5_fold_CV}
\end{table}

\subsubsection{WGAN}\label{subsubsec:WGAN_Experiments}

\begin{figure}[!htb]
\caption{
    The figure shows $G(z)$ and $D(x)$’ training losses for 12 grid search points, where the number of critic training epochs (n\_critic) and batch size are varied, for n\_critic = 2, 5, 10 and 20.}
\centering
\begin{subfigure}[b]{0.47\textwidth}
         \centering
         \includegraphics[width=\textwidth]{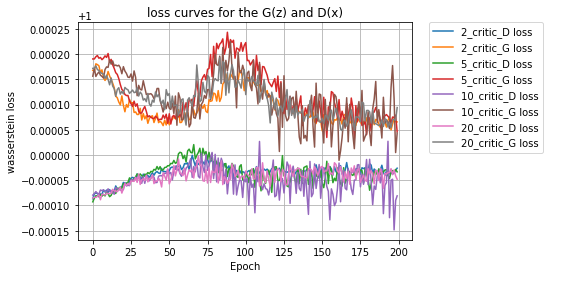}
         \caption{$batch size = 32$}
         \label{fig:wgan1}
     \end{subfigure}
     \hfill
     \begin{subfigure}[b]{0.47\textwidth}
         \centering
         \includegraphics[width=\textwidth]{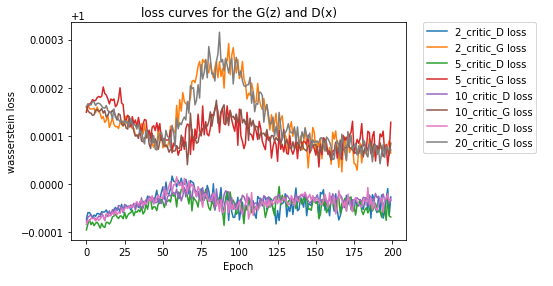}
         \caption{$batch size = 64$}
         \label{fig:wgan2}
     \end{subfigure}
     \hfill
     \begin{subfigure}[b]{0.5\textwidth}
         \centering
         \includegraphics[width=\textwidth]{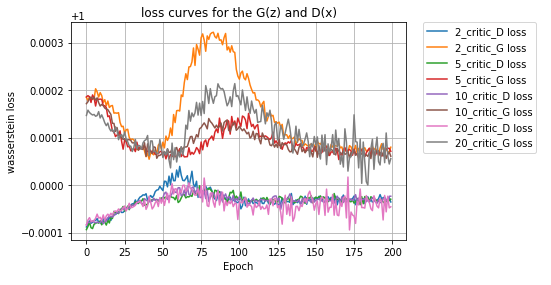}
         \caption{$batch size = 128$}
         \label{fig:wgan3}
     \end{subfigure}
    \label{fig:WGAN_LOSS}
\end{figure}

Table \ref{tab:WGAN_hyper} shows the final training hyper-parameters for both the generator and discriminator LSTM networks of the WGAN.
Training the WGAN has more moving parts than the encoder-decoder model. The initial training parameters were adopted from  \cite{Salimans:Salimans16}, and then a grid search was used to fine tune the hyper-parameters over 200 epochs. Figure \ref{fig:WGAN_LOSS} shows loss curves for both $G(z)$ and $D(z)$ for different points in the hyper-parameter grid search space. The combination of parameters that led to stable training were chosen as the final training parameters shown in Table \ref{tab:WGAN_hyper}. 

\begin{table}
 \begin{center}
 \begin{tabular}{|l|l|}
  \hline
  \bf Hyper-parameter & \bf Final Value \\
  \hline
  \hline
  $G(z)$ Learning Rate & 0.00005 \\
  $G(z)$ Optimizer & RMSprop \\
  $G(z)$ Epochs & 4000 \\
  $G(z)$ Batch size & 32 \\
  $Z$ Latent Distribution & Standard Normal \\
  $D(x)$ Learning Rate & 0.00004 \\
  $D(x) : G(z)$ Epoch Ratio & 5:1 \\
  Gradient Clipping & 0.01 \\
  \hline
 \end{tabular}
\end{center}
 \caption{Training hyper-parameters for the WGAN model.}
 \label{tab:WGAN_hyper}
\end{table}

At the start of training the generator takes Gaussian noise together with the discriminator’s critic of the generated samples. During this initial training the discriminator’s layers are fixed and not trained to ensure the generator learning has started by the time the discriminator is trained to optimality. For each epoch of the generator, the discriminator is trained for five times more epochs, then it is used to critic the generator at its current competence level. Since the RMSprop is used, the learning rates for both the generator and discriminator decay with the number of epochs. This is to force the generator to value learned signal more than temporary noise caused by sudden inconsistent gradient direction changes. Training of the WGAN model, was faster than that of the encoder-decoder model and had fewer combined generator and discriminator trainable weights. This is because there is no transfer of LSTM cell and hidden states between the generator and discriminator unlike there is between the encoder and decoder networks. The WGAN networks are connected only by a scalar loss for information transfer. The generator and discriminator are not trained to optimize accuracy, since the aim is not regenerating any song from the training set, and so the losses are rather similar to the MSE loss. The early stopping criteria for training was set to be when the EM distance between the current generator’s learned distribution and that of the real data stops reducing for any 10 consecutive epochs.

\subsection{Music Generation}\label{subsec:Music_Generation}
For the encoder-decoder model, music is generated one sample at a time at the end of training by priming the decoder network with a short series of notes and predicting the note progression probabilities for the entire song. The generated samples are on average 1.3 minutes long. It was observed that generating melodies that are much longer than 2 minutes resulted in repetitions of the same sequence of notes and sometimes silent spots or complete silence towards the end of the song. For WGAN, samples are generated during training. At the end of each generator training epoch, a sequence of ten priming latent vectors are passed to the generator, and it then generates ten audio samples based on its proficiency at that point in time. This process is repeated until training is complete, resulting in a number of music samples. If training progresses as expected, the samples reflect an increase in compositional skill from epoch one to the last epoch. Only randomly selected samples from the last generator training were collected and written to a MIDI file for evaluation.

\section{Results}\label{sec:Resuts}
A model’s predictive accuracy is a quantitative measure of the number of prediction instances the model estimated correctly divided by the total number of prediction instances as a percentage. In the case of musical notes, accuracy is measured as mean number of notes correctly classified as on or off per time step. The accuracy is measured during both training and validation of the model to ensure it is not over-fitting and is generalizing well. This is measured on the validation set. The table \ref{tab:Train_Test_Accuracies} shows the loss and accuracy scores for the encoder-decoder model together with the WGAN’s EM distance loss, which similarly to MSE loss, has only a lower bound of zero and no upper bound.

\begin{table}
 \begin{center}
 \begin{tabular}{|l|l|l|}
  \hline
   \bf Metric & \bf WGAN LSTM & \bf Encoder-Decoder LSTM \\
  \hline
  \hline
  Training Accuracy & - & 96.45\% \\
  Test Accuracy & - & 96.3\% \\
  Train Entropy Loss  & - & 0.107  \\
  Test Entropy Loss  & - & 0.081  \\
  Generator Loss  & 1.006 & -  \\
  Discriminator Loss  & 0.9995 & -  \\\hline
 \end{tabular}
\end{center}
 \caption{Top-1 Training and test accuracies for the the Seq2Seq model, as well as loss figures for the WGAN.}
 \label{tab:Train_Test_Accuracies}
\end{table}

\noindent
Although the prediction accuracy reported in Table \ref{tab:Train_Test_Accuracies} seems favourably high, it is a bad measure of the generative ability of the encoder-decoder model on the training data. This is because there are more negative than positive prediction instances in the dataset as described in Section \ref{sec:Database}, and shown in Table \ref{tab:Data_Imbalance}. To solve this problem, BAcc that results in a weighted score between the number of prediction classes was used. Since the models trained depend on random initialization of weights, 5-fold CV was performed to get the average performance of the models. Table \ref{tab:CV_balanced_accuracy_endelstm} shows the CV BAcc and loss that are a more realistic measure of the model’s prediction ability. 

\begin{table}
 \begin{center}
 \begin{tabular}{|l|l|l|l|l|}
  \hline
   \bf CV Iteration & \bf Training BAcc & \bf Validation BAcc & \bf Training Loss & \bf Validation Loss \\
  \hline
  \hline
  1 & 74.16\% & 73.84\% & 0.64\% & 0.78\% \\\hline
  2 & 74.27\% & 74.66\% & 0.65\% & 0.30\% \\\hline
  3 & 74.75\% & 74.78\% & 0.29\% & 0.16\% \\\hline
  4 & 74.34\% & 74.24\% & 0.61\% & 0.67\% \\\hline
  5 & 74.61\% & 74.09\% & 0.31\% & 0.46\% \\\hline
  \bf $\mu$ & \bf 74.42\% & \bf 74.32\% & \bf 0.5\% & \bf 0.48\% \\\hline
  \bf $\sigma$ & \bf 0.24\% & \bf 0.39\% & \bf 0.18\% & \bf 0.25\% \\
  \hline
 \end{tabular}
\end{center}
 \caption{Training and five-fold CV balanced top-1 accuracy scores for the encoder-decoder LSTM neural
 network.}
 \label{tab:CV_balanced_accuracy_endelstm}
\end{table}

\begin{table}
 \begin{center}
 \begin{tabular}{|l|l|l|}
  \hline
  \bf CV Iteration & \bf $G(z)$ Loss & \bf $D(z)$ Loss \\
  \hline
  \hline
  1 & 1.000587 & 0.99953 \\\hline
  2 & 1.000384 & 0.99962 \\\hline
  3 & 1.000457 & 0.99959 \\\hline
  4 & 1.000535 & 0.99956 \\\hline
  5 & 1.000561 & 0.99952 \\\hline
  \bf $\mu$ & \bf 1.0005 & \bf 0.9995 \\\hline
  \bf $\sigma$ & \bf 0.00008 & \bf 0.00017 \\\hline
 \end{tabular}
\end{center}
 \caption{Training and five-fold CV EM loss for the WGAN model.}
 \label{tab:WGAN_CV_Loss}
\end{table}

\begin{figure}[!htb]
\centering
\begin{minipage}[b]{0.45\textwidth}
    \includegraphics[width=\textwidth]{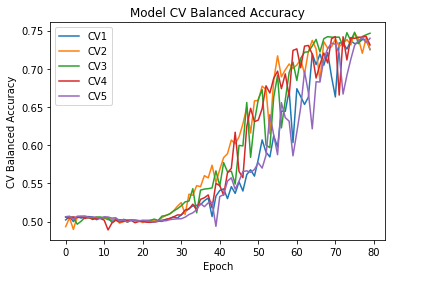}
    \caption{Five-fold CV BAcc curves for the encoder-decoder LSTM.}
    \label{fig:CV_CURVES_ENCODECO}
\end{minipage}
\hfill
\begin{minipage}[b]{0.45\textwidth}
    \includegraphics[width=\textwidth]{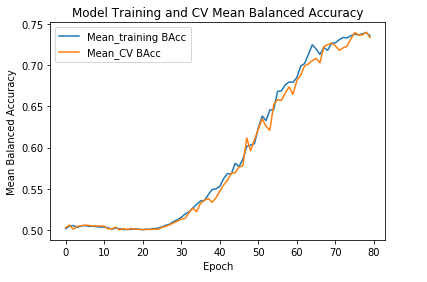}
    \caption{Mean training and CV BAcc curves.}
    \label{fig:CV_CURVES_ENCODECO_1}
\end{minipage}
\end{figure}

Results from Tables \ref{tab:Train_Test_Accuracies}, \ref{tab:CV_balanced_accuracy_endelstm}, and \ref{tab:WGAN_CV_Loss} are not sufficient to arrive at a conclusion in comparing WGAN to the encoder-decoder model for the purpose of music generation, since they follow very different training methods with completely different objective functions. Also, music quality is more subjective as an art form than it is objective, and so there is a lack of good objective measures to base a conclusion on \cite{Mogren:CRNNGAN16,Colombo:BachProp18,Roberts:HeiracicalLVM18}. Accuracy reported for the encoder-decoder model represents the mean fraction of correct pitch predictions made by the decoder in the training song per prediction class, and is depicted in the training curves shown in Figure \ref{fig:CV_CURVES_ENCODECO_1}. The EM distance for WGAN is the final distance of the generators learned sampling distribution from that of the training data distribution.

\begin{figure}[!htb]
\caption{Loss curves for WGAN with the best grid search parameters: n\_critic=5 and batch size=32.}
\label{fig:WGAN_LOSS_CURVES}
\centering
\includegraphics[scale=0.8]{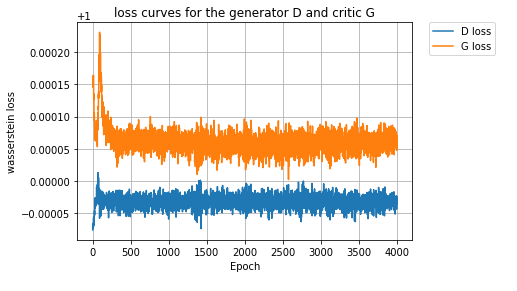}
\end{figure}

The close training and CV mean BAcc curves in Figure \ref{fig:CV_CURVES_ENCODECO} also show the encoder-decoder LSMT is not over-fitting. An over-fitting model would regenerate the training music data samples when primed with a close enough sequence of notes and would lack diversity in the generated samples. Although the encoder-decoder does not seem to be over-fitting based on the CV accuracy, volunteers in the listening survey do hint at lack of diversity and creativity in the encoder-decoder’s generated samples, discussed in Section \ref{subsec:Listener_Comments}.  Figure \ref{fig:WGAN_LOSS_CURVES} contains training EM distance loss curves for the WGAN networks. Note that the generator is much more unstable than the discriminator. The generator produces an estimate of the true distribution, and the discriminator’s loss estimates how far off the generator is through the EM distance. As stated earlier, due to the subjective nature of music, a subjective evaluation method was used for the generated samples and the results are presented in Section \ref{subsec:MOS}

\subsection{Mean Opinion Scores}\label{subsec:MOS}
\noindent
Tables \ref{tab:MOS_Scores_WGAN} and \ref{tab:MOS_Scores_Encoder_Decoder} present the results of the listener impression survey for the two generative models.

\begin{table}
 \begin{center}
 \begin{tabular}{|l|l|l|l|l|l|l|l|l|}
  \hline
   \bf Volunteer & \bf S1 & \bf S2 & \bf S3 & \bf S4 & \bf S5 & \bf S6 & \bf S7 & \bf S8\\
  \hline
  \hline
  1 & 2.5 & 4 & 2 & 3 & 2.5 & 4 & 4 & 3\\
  2 & 4 & 3.6 & 4 & 3 & 3 & 2 & 3.5 & 2.5\\
  3 & 3 & 3 & 3 & 4 & 3 & 3.5 & 4 & 3\\
  4 & 4 & 4 & 2 & 3 & 3.5 & 4 & 3 & 2\\
  5 & 3.5 & 4 & 4 & 4 & 3 & 2.5 & 3 & 4\\
  6 & 3 & 1 & 2 & 4 & 2 & 3 & 3.5 & 2.5\\
  7 & 2.8 & 4 & 5 & 3.5 & 1 & 5 & 3 & 3\\
  8 & 2.5 & 4 & 3.5 & 3 & 3 & 4 & 4 & 3\\
  9 & 1.5 & 4 & 2.5 & 3 & 4 & 4 & 3 & 4.5\\
  10 & 2 & 4 & 2.5 & 4 & 3 & 3 & 3.5 & 2.5\\ \hline
  \bf $q(s_{i})$ & 2.88 & 3.56 & 3.05 & 3.45 & 2.88 & 3.5 & 3.45 & 3\\ \hline
  \bf $\sigma_{q}(s_{i})$ & 0.77 & 0.91 & 0.99 & 0.47 & 0.78 & 0.84 & 0.42 & 0.71\\ \hline
 \end{tabular}
\end{center}
 \caption{Listener impression scores for WGAN generated samples S1 to S8. $q(s_{i})$ represents the
MOS for each sample}
 \label{tab:MOS_Scores_WGAN}
\end{table}

\begin{table}
 \begin{center}
 \begin{tabular}{|l|l|l|l|l|l|l|l|l|}
  \hline
  \bf Volunteer & \bf S9 & \bf S10 & \bf S11 & \bf S12 & \bf S13 & \bf S14 & \bf S15 & \bf S16\\
  \hline
  \hline
  1 & 3 & 3.3 & 1 & 2 & 2 & 3 & 2.5 & 3.5\\
  2 & 2 & 3 & 3.5 & 2 & 3 & 3 & 4 & 2\\
  3 & 2 & 4 & 2.5 & 2 & 2 & 3.5 & 3 & 2\\
  4 & 4.5 & 3 & 3 & 5 & 2 & 3 & 2.5 & 3\\
  5 & 2 & 2 & 3 & 2 & 1 & 3 & 2 & 2\\
  6 & 4 & 2 & 3 & 3 & 3.5 & 2.5 & 4 & 3.5\\
  7 & 2.5 & 3 & 4 & 2 & 3 & 4 & 2 & 4\\
  8 & 3.5 & 3 & 3 & 3.5 & 2 & 3.5 & 3 & 2\\
  9 & 3 & 2 & 3.5 & 2 & 3 & 3 & 3 & 3\\
  10 & 3.5 & 3 & 3.5 & 3 & 3 & 4 & 2 & 4\\ \hline
  \bf $q(s_{i})$ & 3 & 2.83 & 3 & 2.65 & 2.45 & 3.25 & 2.8 & 2.9\\ \hline
  \bf $\sigma_{q}(s_{i})$ & 0.84 & 0.62 & 0.77 & 0.95 & 0.72 & 0.46 & 0.71 & 0.8\\ \hline
 \end{tabular}
\end{center}
 \caption{Listener impression scores for the encoder-decoder LSTM generated samples S9 to S16.}
 \label{tab:MOS_Scores_Encoder_Decoder}
\end{table}

\begin{table}
 \begin{center}
 \begin{tabular}{|l|l|l|}
  \hline
    & \bf WGAN LSTM & \bf Encoder-Decoder LSTM \\
  \hline
  \hline
  $Q_{q_{s}}$ & 3.21 & 2.86 \\
  $\sigma_{Q_{q_{s}}}$ & 0.811 & 0.736 \\
  $Q_{q_{s}} + 2\times \sigma_{Q_{q_{s}}}$ & 4.832 & 4.332 \\
  $Q_{q_{s}} - 2\times \sigma_{Q_{q_{s}}}$ & 1.588 & 1.388 \\
  Median Score & 3 & 3 \\\hline
 \end{tabular}
\end{center}
 \caption{Mean and Median generator quality scores. Although the sample median scores from the two models are equal, this does not imply they are drawn from populations with equal median scores. It is the Wilcoxon signed-rank t-test that gives a conclusive answer on equality of the population medians.}
 \label{tab:Mean_And_Median_Qs}
\end{table}

\noindent
The results in Tables \ref{tab:MOS_Scores_WGAN} and \ref{tab:MOS_Scores_Encoder_Decoder} show that volunteers generally scored WGAN samples higher than the encoder-decoder samples. This is reflected in the observation that only one of four WGAN samples received a MOS below three, yet all encoder-decoder samples got a MOS of three or lower. All samples from both models had score variations that are considerably low (all below one), also, both models have 95\% sample ratings that are within two standard deviations of each other. This is a good indication that as much as the rating system is subjective and high variance is expected, people’s opinions about the samples do not differ so significantly that it were to seem they are each receiving a song of a different genre or were all exposed to entirely different interpretations of what good music sounds like.

The overall results for both models across all samples are presented in Table \ref{tab:Mean_And_Median_Qs}, and are be used to deduce which of the two models is considered a more skilled composer of music than the other based on its generated samples. The results in Table \ref{tab:Mean_And_Median_Qs} show that the WGAN samples have a higher mean opinion score than those of the encoder-decoder model based on volunteer listener’s ratings. However, since the arithmetic mean is a measure that can be greatly influenced by outliers, it is also important to consider the median opinion scores which is not influenced by outliers, and is a measure of centrality of data observations. The Wilcoxon signed-rank test is used to assess whether the opinion scores collected for the WGAN and encoder-decoder models come from populations with equal medians. The Wilcoxon test results are discussed in the following section.

\subsection{Wilcoxon Signed-Rank Test}\label{subsec:Wilcoxon}
\begin{figure}[!htb]
\caption{Opinion score distribution for WGAN and encoder-decoder LSTM generated music samples.}
\label{fig:OPINION_SCORE_FREQUENCY}
\centering
\includegraphics[scale=0.9]{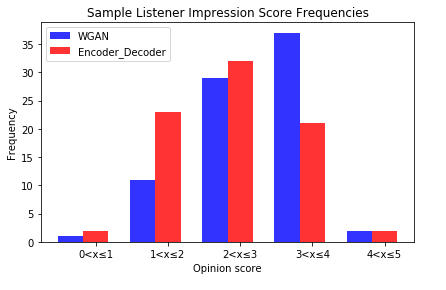}
\end{figure}

The Wilcoxon signed-rank test for equal population medians as explained in Section \ref{subsec: Wilcoxon Signed-Rank T-Test} was performed. Using Equations \ref{eq:wilcoxon_statistic}, \ref{eq:z_score} and \ref{eq:sigmaW}, the test statistic $W$, signed-rank standard deviation, $\sigma_{W}$, and the $z-score: z$, were calculated and are presented In Table \ref{tab:Wilcoxon_signed_rank_results}. The results are used in making the decision to reject, or to not reject the null hypothesis of equal population medians. It is standard to perform hypothesis tests to a certain level of confidence, usually $\alpha = 0.95$. The critical $z_{0.95}$ value is the inverse standard normal value under which 95\% of all data falls, since the standardized signed-rank test statistic $z$ follows a standard normal distribution. Results in Table \ref{tab:Wilcoxon_signed_rank_results} show that $z > z_{0,95}$ and according to the Wilcoxon signed-rank test, the null hypothesis of equal medians is rejected with 95\% confidence in favour of the alternative hypothesis. The alternative hypothesis states the two generative models produced samples with significantly different opinion score medians, and thus their population distributions are not centered around the same score. This is also supported by plotting a histogram of opinion scores for both models in Figure \ref{fig:OPINION_SCORE_FREQUENCY}.

\begin{table}
 \begin{center}
 \begin{tabular}{|l|l|l|l|}
  \hline
    \bf W & \bf $\sigma_{W}$ & \bf $|Z|$ & \bf $Z_{\alpha = 0.95}$ \\
  \hline
  \hline
  -1132 & 393.90 & 2.8751 &1.6449\\
  \hline
 \end{tabular}
\end{center}
 \caption{Wilcoxon signed-rank test results.}
 \label{tab:Wilcoxon_signed_rank_results}
\end{table}

Figure \ref{fig:OPINION_SCORE_FREQUENCY} shows that WGAN sample scores are skewed more to the right as compared to those of the encoder-decoder network that form a more symmetric distribution. The evident skewed WGAN sample score distribution and a higher median opinion score suggest the higher WGAN MOS score is not falsely influenced by outliers, and this too supports that WGAN samples were found to be more pleasing to listen to than those generated by the encoder-decoder model.

% Acknowledgements should go at the end, before appendices and references

\subsection{Listener Comments}\label{subsec:Listener_Comments}
Although volunteers were not asked to provide any textual comments, some of them did provide written feedback together with the rating scores. Majority of the comments provided justification for the volunteer’s rating of the samples and their preference for one sample over the other. The generating model identity of the samples were not known to the volunteers during the survey, named as simply sample 1 to 8 for the WGAN, and 9 to 16 for the encoder-decoder model. In total, 26 of the ratings were accompanied by a comment. Some of the most insightful comments are provided in appendix \ref{appendix:a}, with the generating model’s name instead of the sample number revealed. The comments point to a general preference for WGAN music samples over those of the encoder-decoder model. WGAN sample comments can be summarized using the following keywords: melodic, rhythmic, nice, creative. The encoder-decoder samples can be described: Slow, vague, rhythmic.

\section{Conclusion}\label{sec:Conclusion}
This work described and compared two training configurations of generative models for music generation using MIDI data. After a detailed discussion of neural network architectures normally used for time series data and their topological components, state of the art music composition networks were discussed. A short background on the MIDI file format and its representation in this work was provided. The work then detailed the WGAN and encoder-decoder models with LSTM components implemented and compared for the composition task, followed by the experimental setup and results. The main purpose of the study was to first show that the adversarial configuration is a viable approach to training neural networks for music composition, and that it produces music samples of superior quality as compared to non-adversarial training. To be able to evaluate the generative skill of the two compared training configurations, a survey was conducted where 10 volunteers were asked to listen to four randomly selected samples from each generative model, and rate each sample on a scale from 1 to 5 based on how pleasing it was to listen to. 70\% of rating instances resulted in preference for WGAN over the encoder-decoder samples. 30\% of volunteer ratings pointed to a preference for encoder-decoder generated samples over WGAN samples. The ratings were then used to conduct a rating test to determine whether the two model’s sample ratings were from populations with different medians, and this test concluded they were. Assuming the sample ratings are representative enough of their populations, this implies adding more volunteers to the survey would infer that the adversarial neural network is a better composer than the encoder-decoder neural network. Based on the results of the survey and comments from volunteers on the generated samples, it can be concluded that adversarial training is a viable method for training generative models for music generation, and it does produce more diverse and pleasing to listen to music samples. Further experiments can be conducted on the WGAN implementation above by including more varied training music genres, and allowing the generator to generate longer sequences. Since the generator is an LSTM network, methods normally used to improve performance and memory retention in sequence-to-sequence models on natural language processing (NLP) tasks, such as adding an attention \cite{See:Attention17} or pointer \cite{Vinyals:Pointers15} layer can be explored. It would be interesting as well to perform a thorough investigation of pure transformer networks \cite{Vaswani:AttentionIsAll17} in music generation, as well as how training methods such as diffusion \cite{Jonathan:DenoisingDiffusion20,Mittal:SymbolicDiffusionMusic21,Andreas:RepaintDiffusion22} can be applied to music. Another avenue of further research to be explored is the development and consideration of  formal and objective evaluation methods for artistic tasks such as music generation and artistic painting. Being able to accurately and appropriately evaluate models for such subjective tasks will greatly improve the quality of generated samples.


\appendix
\section{Appendix :Listener comments}\label{appendix:a}
\begin{itemize}
  \item ``I couldn’t get a feel of where the encoder-decoder song is going, the WGAN sample has a nice classical feel to it."
  \item ``They both sound musical but the sound quality is bad."
  \item ``I like the pace of the WGAN sample."
  \item ``The encoder-decoder sample had too many silent spaces, the quiet spots emphasize the loud spots ,sounded like rambling."
  \item ``The WGAN sample is a little chaotic but generally creates a good atmosphere, the encoder-decoder song has good rhythm but no actual melody.
  \item ``I’d say the encoder-decoder song is better, has more rhythm, the WGAN sample sounds too fast and just noise to me."
  \item ``The WGAN Sample sounds like me when I am under dissertation stress, terrible."
  \item ``The WGAN sample sounds more creative, the encoder-decoder sample wins on rhythm although it takes soo long to get there."
  \item ``What’s important to me is they all sound musical."
  \item ``I don’t listen to this genre so my view may be way off."
  \item ``Most of the songs sound similar to me."
  \item ``Wow, can’t believe the WGAN sample was generated by a machine, though it’s funny at the end LOL."
  \item ``Can’t you get them to generate longer songs? perhaps with words? I like the encoder-decoder sample."
\end{itemize}

\bibliographystyle{unsrtnat}
\bibliography{references}

\begin{thebibliography}{57}
\providecommand{\natexlab}[1]{#1}
\providecommand{\url}[1]{\texttt{#1}}
\expandafter\ifx\csname urlstyle\endcsname\relax
  \providecommand{\doi}[1]{doi: #1}\else
  \providecommand{\doi}{doi: \begingroup \urlstyle{rm}\Url}\fi

\bibitem[Sutskever et~al.(2014)Sutskever, Vinyals, and Le]{Sutskever:Seq2Seq14}
I.~Sutskever, O.~Vinyals, and Q.~Le.
\newblock Sequence to sequence learning with neural networks.
\newblock In \emph{Advances in neural information processing systems},
  volume~4, 2014.

\bibitem[Liu and Randall(2016)]{Liu:MissingMusic16}
I.~Liu and R.~Randall.
\newblock Predicting missing music components with bidirectional long
  short-term memory neural networks.
\newblock In \emph{Proceedings of the 17th International Society on Music
  Information Retrieval Conference}, 2016.

\bibitem[Yang et~al.(2017)Yang, Chuo, and Yang]{Yang:MidiNet17}
L.~Yang, S.~Chuo, and Y.~Yang.
\newblock Midinet: A convolutional generative adversarial network for
  symbolic-domain music generation.
\newblock In \emph{Proceedings of the 18th International Society on Music
  Information Retrieval Conference}, pages 1--12, 2017.

\bibitem[Dong et~al.(2018)Dong, Hsiao, Yang, and Yang]{Dong:MuseGan18}
H.~Dong, W.~Hsiao, L.~Yang, and Y.~Yang.
\newblock Musegan: Multi-track sequential generative adversarial networks for
  symbolic music generation and accompaniment.
\newblock In \emph{Proceedings of the the 30th international conference on
  Innovative Applications of Artificial Intelligence}, pages 34--41, 2018.

\bibitem[Chung et~al.(2014)Chung, Gulcehre, Cho, and
  Bengio]{Chung:EmpericalRNN14}
J.~Chung, C.~Gulcehre, K.~Cho, and Y.~Bengio.
\newblock Empirical evaluation of gated recurrent neural networks on sequence
  modeling.
\newblock In \emph{Proceedings of the 27th international conference on Neural
  Information Processing Systems, Deep Learning and Representation Learning},
  2014.

\bibitem[Waite(2016)]{Waite:LookBackRNN2016}
E.~Waite.
\newblock Generating long-term structure in songs and stories.
\newblock In \emph{lookback-rnn-attention-rnn}, 2016.

\bibitem[Weel(2017)]{Weel:RoboMozart17}
J.~Weel.
\newblock Robomozart:generating music using lstm networks trained per-tick on a
  midi collection with short music segments as input, 2017.

\bibitem[Oord et~al.(2016{\natexlab{a}})Oord, Dieleman, Zen, Simonyan, Vinyals,
  Graves, Kalchbrenner, Senior, and Kavukcuoglu]{Oord:WaveNet16}
A.~Oord, S.~Dieleman, H.~Zen, K.~Simonyan, O.~Vinyals, A.~Graves,
  N.~Kalchbrenner, A.~Senior, and K.~Kavukcuoglu.
\newblock Wavenet: A generative model for raw audio.
\newblock In \emph{Proceedings of the 9th International Speech Communication
  Association, Speech Synthesis Workshop}, pages 125--125, 2016{\natexlab{a}}.

\bibitem[Mogren(2016)]{Mogren:CRNNGAN16}
O.~Mogren.
\newblock C-rnn-gan: Continuous recurrent neural networks with adversarial
  training.
\newblock In \emph{Proceedings of the 30th International Conference on Neural
  Information Processing Systems, Constructive Machine Learning Workshop},
  2016.

\bibitem[Lerdahl and Jackendoff(1983)]{Lerdahl:TonalGen83}
F.~Lerdahl and R.~Jackendoff.
\newblock On the algorithmic description of the process of composing music.
\newblock In \emph{MIT Press}, 1983.

\bibitem[Colombo and Gerstner(2018)]{Colombo:BachProp18}
F.~Colombo and W.~Gerstner.
\newblock Bachprop:learning to compose music in multiple styles.
\newblock In \emph{Arxiv}, volume abs/1802.05162, 2018.

\bibitem[Cho et~al.(2014)Cho, Merrienboer, Gulcehre, Bahdanau, Bougares,
  Schwenk, and Bengio]{Cho:EnDe14}
K.~Cho, B.~Merrienboer, C.~Gulcehre, D.~Bahdanau, F.~Bougares, H.~Schwenk, and
  Y.~Bengio.
\newblock Learning phrase representations using rnn encoder–decoder for
  statistical machine translation.
\newblock In \emph{Proceedings of the 2014 Conference on Empirical Methods in
  Natural Language Processing}, pages 1724--1734, 2014.

\bibitem[Arjovsky et~al.(2017)Arjovsky, Chintala, and Bottou]{Arjovsky:WGAN17}
M.~Arjovsky, S.~Chintala, and L.~Bottou.
\newblock Wasserstein gan.
\newblock In \emph{Proceedings of the 34th International Conference on Machine
  Learning}, volume~70, pages 214--223, 2017.

\bibitem[Shiebler(2017)]{Shiebler:MusicRNNRBM17}
D.~Shiebler.
\newblock Music rnn rbm.
\newblock In \emph{github.com/dshieble/Music\_RNN\_RBM}, 2017.

\bibitem[Zaripov(1960)]{Zaripov:AlgoMusic60}
R.~Zaripov.
\newblock On the algorithmic description of the process of composing music.
\newblock In \emph{In USSR Academy of Sciences}, volume 132, pages 1283--1286,
  1960.

\bibitem[Xenakis(1992)]{Xenakis:FormalMusic92}
I.~Xenakis.
\newblock Formalized music: Thought and mathematics in music.
\newblock In \emph{revised ed. Pendragon}, 1992.

\bibitem[Alfonseca et~al.(2007)Alfonseca, Cebrian, and
  Puente]{Alfonseca:Genetic07}
M.~Alfonseca, M.~Cebrian, and O.~Puente.
\newblock A simple genetic algorithm for music generation by means of
  algorithmic information theory.
\newblock In \emph{Proceedings of the Institute of Electrical and Electronics
  Engineers Congress, Evolutionary Computation}, pages 25--28, 2007.

\bibitem[Hochreiter and Schmidhuber(1997)]{Hochreiter:LSTM97}
S.~Hochreiter and J.~Schmidhuber.
\newblock Long short-term memory.
\newblock In \emph{Neural Computation}, volume~9, pages 1735--1780, 1997.

\bibitem[Goodfellow et~al.(2014)Goodfellow, Pouget-Abadie, Mirza, Xu,
  Warde-Farley, Ozair, Courville, and Bengio]{Goodfellow:GANs14}
I.~Goodfellow, J.~Pouget-Abadie, M.~Mirza, B.~Xu, D.~Warde-Farley, S.~Ozair,
  A.~Courville, and Y.~Bengio.
\newblock Generative adversarial nets.
\newblock In \emph{Proceedings of the 27th International Conference on Neural
  Information Processing Systems}, pages 2672--2680, 2014.

\bibitem[Elgammal et~al.(2017)Elgammal, Liu, Elhoseiny, and
  Mazzone]{Elgammal:CAN17}
A.~Elgammal, B.~Liu, M.~Elhoseiny, and M.~Mazzone.
\newblock Can: Creative adversarial networks generating "art" by learning
  styles and deviating from style norms.
\newblock In \emph{Proceedings of the International Conference on Computational
  Creativity}, 2017.

\bibitem[Juefei-Xu et~al.(2017)Juefei-Xu, Boddeti, and
  Savvides]{Juefei-Xu:GangOG17}
F.~Juefei-Xu, V.~N. Boddeti, and M.~Savvides.
\newblock Gang of gans:generative adversarial networks with maximum margin
  ranking.
\newblock In \emph{ArXiv}, volume abs/1704.04865, 2017.

\bibitem[Yu et~al.(2017)Yu, Zhang, Wang, and Yu]{YU:SeqGAN17}
L.~Yu, W.~Zhang, J.~Wang, and Y.~Yu.
\newblock Seqgan: Sequence generative adversarial nets with policy gradient.
\newblock In \emph{Proceedings of the 31st AAAI Conference on Artificial
  Intelligence}, volume~31, 2017.

\bibitem[Hung et~al.(2021)Hung, Chen, Yeh, and
  Yang]{Hung:FreesoundGeneration21}
T.~Hung, B.~Chen, Y.~Yeh, and Y.~Yang.
\newblock A benchmarking initiative for audio-domain music generation using the
  freesound loop dataset.
\newblock In \emph{Proceedings of the 22nd International Society for Music
  Information Retrieval Conference, {ISMIR} 2021, Online, November 7-12, 2021},
  pages 310--317, 2021.
\newblock URL \url{https://archives.ismir.net/ismir2021/paper/000038.pdf}.

\bibitem[António et~al.(2020)António, Frederic, Dmitry, Jordan, Hsuan, Joann,
  Yu, Kao, Hsu, and Xavier]{António:FreeLoops20}
R.~António, F.~Frederic, B.~Dmitry, S.~Jordan, Y.~Hsuan, C.~Joann, C.~Yu,
  W.~Kao, W.~Hsu, and S.~Xavier.
\newblock The freesound loop dataset and annotation tool.
\newblock In \emph{arXiv}, 08 2020.

\bibitem[Denil et~al.(2012)Denil, Bazzani, Larochelle, and
  de~Freitas]{Denil:Attend12}
M.~Denil, L.~Bazzani, H.~Larochelle, and N.~de~Freitas.
\newblock Learning where to attend with deep architectures for image tracking.
\newblock In \emph{Institute of Electrical and Electronics Engineers, Neural
  Computation}, volume~24, pages 2151--2184, 2012.

\bibitem[Oord et~al.(2016{\natexlab{b}})Oord, Kalchbrenner, and
  Kavukcuoglu]{Oord:PixelNet16}
A.~Oord, N.~Kalchbrenner, and K.~Kavukcuoglu.
\newblock Pixel recurrent neural networks.
\newblock In \emph{Proceedings of the 33rd International Conference on Machine
  Learning}, volume~48, pages 1747--1756, 2016{\natexlab{b}}.

\bibitem[Chen et~al.(2022)Chen, Hsu, Liao, Ram{\'{\i}}rez, Mitsufuji, and
  Yang]{Chen:AutomaticGANDJ22}
B.~Chen, E.~Hsu, W.~Liao, M.~Ram{\'{\i}}rez, Y.~Mitsufuji, and Y.~Yang.
\newblock Automatic {DJ} transitions with differentiable audio effects and
  generative adversarial networks.
\newblock In \emph{{IEEE} International Conference on Acoustics, Speech and
  Signal Processing, {ICASSP} 2022, Virtual and Singapore, 23-27 May 2022},
  pages 466--470. {IEEE}, 2022.
\newblock \doi{10.1109/ICASSP43922.2022.9746663}.
\newblock URL \url{https://doi.org/10.1109/ICASSP43922.2022.9746663}.

\bibitem[Edirisooriya et~al.(2021)Edirisooriya, Dong, McAuley, and
  Berg-Kirkpatrick]{Edirisooriya:OMR21}
S.~Edirisooriya, H.W. Dong, J.~McAuley, and T.~Berg-Kirkpatrick.
\newblock An empirical evaluation of end-to-end polyphonic optical music
  recognition.
\newblock In \emph{International Society for Music Information Retrieval},
  2021.

\bibitem[Chu et~al.(2017)Chu, Urtasun, and Fidler]{Chu:PiMusic17}
H.~Chu, R.~Urtasun, and S.~Fidler.
\newblock Song from pi: A musically plausible network for pop music generation.
\newblock In \emph{Proceedings of the 5th International Conference on Learning
  Representations}, 2017.

\bibitem[Alexey and Ivan(2020)]{Alexey:RVAEHistory20}
T.~Alexey and Y.~Ivan.
\newblock Music generation with variational recurrent autoencoder supported by
  history.
\newblock \emph{SN Applied Sciences}, 2, 12 2020.
\newblock \doi{10.1007/s42452-020-03715-w}.

\bibitem[Roberts et~al.(2017)Roberts, Engel, and Eck]{Roberts:HVAEMusic17}
A.~Roberts, J.~Engel, and D.~Eck.
\newblock Hierarchical variational autoencoders for music.
\newblock In \emph{Workshop on Machine Learning for Creativity and Design,
  NIPS}, 2017.

\bibitem[Vincent et~al.(2012)Vincent, Lewandowski, and
  Bengio]{Vincent:TemporalDependencies12}
P.~Vincent, N.~Lewandowski, and Y.~Bengio.
\newblock Modeling temporal dependencies in high-dimensional
  sequences:application to polyphonic music generation and transcription.
\newblock In \emph{Proceedings of the 27th International Conference on Machine
  Learning}, 2012.

\bibitem[Bretan et~al.(2017)Bretan, Weinberg, and Heck]{Bretan:UnitSelection17}
M.~Bretan, G.~Weinberg, and L.~Heck.
\newblock A unit selection methodology for music generation using deep neural
  networks.
\newblock In \emph{Proceedings of the International Conference on Computational
  Creativity}, 2017.

\bibitem[Shuqi et~al.(2021)Shuqi, Zeyu, Celso, and
  Roger]{Shuqi:DaiControllable21}
D.~Shuqi, J.~Zeyu, G.~Celso, and D.~Roger.
\newblock Controllable deep melody generation via hierarchical music structure
  representation.
\newblock \emph{arXiv preprint arXiv:2109.00663}, 2021.

\bibitem[Vaswani et~al.(2017)Vaswani, N.Shazeer, Parmar, Uszkoreit, Jones,
  Gomez, Kaiser, and Polosukhin]{Vaswani:AttentionIsAll17}
A.~Vaswani, N.Shazeer, N.~Parmar, J.~Uszkoreit, L.~Jones, A.~Gomez, L.~Kaiser,
  and I.~Polosukhin.
\newblock Attention is all you need.
\newblock In \emph{Proceedings of the 31st International Conference on Neural
  Information Processing Systems}, pages 6000–--6010, 2017.
\newblock URL \url{https://arxiv.org/pdf/1706.03762.pdf}.

\bibitem[Raffel(2018)]{Raffel:LearningBasedMethods18}
C.~Raffel.
\newblock Learning-based methods for comparing sequences, with applications to
  audio-to-midi alignment and matching.
\newblock In \emph{https://colinraffel.com/projects/lmd/}, 2018.

\bibitem[Henning et~al.(2010)Henning, Soon, Enno, and
  Joachim]{Henning:BalancedAccuracy10}
B.~Henning, O.~Soon, S.~Enno, and B.M. Joachim.
\newblock The balanced accuracy and its posterior distribution.
\newblock In \emph{20th International Conference on Pattern Recognition}, pages
  3121--3124, 2010.
\newblock \doi{10.1109/ICPR.2010.764}.

\bibitem[Vicente et~al.(2009)Vicente, Mollineda., and
  José]{Vicente:Indexbalanced09}
G.~Vicente, R.~Mollineda., and S.~José.
\newblock Index of balanced accuracy: A performance measure for skewed class
  distributions.
\newblock In \emph{Proceedings of the 4th Iberian Conference on Pattern
  Recognition and Image Analysis}, volume 5524, pages 441--448, 06 2009.

\bibitem[Chongsheng et~al.(2022)Chongsheng, Paolo, Jingjun, Gaojuan, George,
  Salvador, and Weiping]{Chongsheng:ImbalancedFeature22}
Z.~Chongsheng, S.~Paolo, B.~Jingjun, F.~Gaojuan, A.~George, G.~Salvador, and
  D.~Weiping.
\newblock An empirical study on the joint impact of feature selection and data
  resampling on imbalance classification.
\newblock In \emph{Applied Intelligence}, pages 1--13, 06 2022.
\newblock \doi{10.1007/s10489-022-03772-1}.

\bibitem[Werbos(1990)]{Werbos:BPTT90}
P.~Werbos.
\newblock Backpropagation through time: what it does and how to do it.
\newblock In \emph{Proceedings of the Institute of Electrical and Electronics
  Engineers}, volume~78, pages 1550--1560, 1990.

\bibitem[Cui et~al.(2017)Cui, Ke, and Wang]{Cui:StackedLSTM17}
Z.~Cui, R.~Ke, and Y.~Wang.
\newblock Deep stacked bidirectional and unidirectional lstm recurrent neural
  network for network-wide traffic speed prediction.
\newblock In \emph{6th International Workshop on Urban Computing}, 2017.

\bibitem[Schuster and Paliwal(1997)]{Schuster:BiLSTM97}
M.~Schuster and K.~Paliwal.
\newblock Bidirectional recurrent neural networks.
\newblock In \emph{Institute of Electrical and Electronics Engineers, Signal
  Processing}, volume~45, pages 2673--2681, 1997.

\bibitem[Mukkamala and Hein(2017)]{Mukkamala:RMSProp17}
M.~C. Mukkamala and M.~Hein.
\newblock Variants of rmsprop and adagrad with logarithmic regret bounds.
\newblock In \emph{Proceedings of the 34th International Conference on Machine
  Learning}, volume~70, pages 2545--2553, 2017.

\bibitem[Lopyrev(2015)]{Lopyrev:15}
K.~Lopyrev.
\newblock Generating news headlines with recurrent neural networks.
\newblock In \emph{ArXiv}, volume abs/1512.01712, 2015.

\bibitem[Kim(2017)]{Kim:KimDeepJaz17}
J.~Kim.
\newblock Using keras and theano for deep learning driven jazz generation.
\newblock In \emph{https://deepjazz.io}, 2017.

\bibitem[Roberts et~al.(2018)Roberts, Engel, Raffel, Hawthorne, and
  Eck]{Roberts:HeiracicalLVM18}
A.~Roberts, J.~Engel, C.~Raffel, C.~Hawthorne, and D.~Eck.
\newblock A hierarchical latent vector model for learning long-term structure
  in music.
\newblock In \emph{Proceedings of the 35th International Conference on Machine
  Learning, Machine Learning Research}, volume~80, pages 4364,4373, 2018.

\bibitem[Hui(2018)]{Hui:WGANGP18}
J.~Hui.
\newblock Gan-wasserstein gan and wgan-gp.
\newblock In \emph{medium}, 2018.

\bibitem[Lackner(2016)]{Lackner:MelodyLSTM16}
K.~Lackner.
\newblock Composing a melody with long-short term memory (lstm) recurrent
  neural networks.
\newblock In \emph{Bachelor’s thesis, Technische Universität München,
  Munich, Germany}, 2016.

\bibitem[Huang and Wu(2016)]{Huang:DL4Music16}
A.~Huang and R.~Wu.
\newblock Deep learning for music.
\newblock In \emph{Arxiv}, volume abs/1606.04930, 2016.

\bibitem[Wilcoxon(1945)]{Wilcoxon:PiMusic17}
F.~Wilcoxon.
\newblock Individual comparisons by ranking methods.
\newblock In \emph{Biometrics Bulletin}, volume~1, pages 80--83, 1945.

\bibitem[Srivastava et~al.(2014)Srivastava, Hinton, Krizhevsky, Sutskever, and
  Salakhutdinov]{Srivastava:Dropout14}
N.~Srivastava, G.E. Hinton, A.~Krizhevsky, I.~Sutskever, and R.~Salakhutdinov.
\newblock Dropout: A simple way to prevent neural networks from overfitting.
\newblock \emph{Journal of Machine Learning Research}, 15\penalty0
  (1):\penalty0 1929--1958, 2014.

\bibitem[Salimans et~al.(2016)Salimans, Goodfellow, Zaremba, Cheung, Radford,
  and Chen]{Salimans:Salimans16}
T.~Salimans, I.~Goodfellow, W.~Zaremba, V.~Cheung, A.~Radford, and X.~Chen.
\newblock Improved techniques for training gans.
\newblock In \emph{Proceedings of the 30th International Conference on Neural
  Information Processing Systems}, pages 2234--2242, 2016.

\bibitem[See et~al.(2017)See, Liu, and Manning]{See:Attention17}
A.~See, P.~Liu, and C.~Manning.
\newblock Get to the point: Summarization with pointer-generator networks.
\newblock In \emph{Proceedings of the 55th Annual Meeting of the Association
  for Computational Linguistics}, volume~1, pages 1073--1083, 2017.

\bibitem[Vinyals et~al.(2015)Vinyals, Fortunato, and
  Jaitly]{Vinyals:Pointers15}
O.~Vinyals, M.~Fortunato, and N.~Jaitly.
\newblock Pointer networks.
\newblock In \emph{Proceedings of the 28th International Conference on Neural
  Information Processing Systems}, volume~2, pages 2692--2700, 2015.

\bibitem[Jonathan et~al.(2020)Jonathan, Ajay, and
  Pieter]{Jonathan:DenoisingDiffusion20}
H.~Jonathan, J.~Ajay, and A.~Pieter.
\newblock Denoising diffusion probabilistic models.
\newblock In \emph{Advances in Neural Information Processing Systems},
  volume~33, pages 6840--6851. Curran Associates, Inc., 2020.
\newblock URL
  \url{https://proceedings.neurips.cc/paper/2020/file/4c5bcfec8584af0d967f1ab10179ca4b-Paper.pdf}.

\bibitem[Mittal et~al.(2021)Mittal, Engel, Hawthorne, and
  Simon]{Mittal:SymbolicDiffusionMusic21}
G.~Mittal, J.~Engel, C.~Hawthorne, and I.~Simon.
\newblock Symbolic music generation with diffusion models.
\newblock In \emph{22nd Proceedings of the International Society for Music
  Information Retrieval}, 2021.
\newblock URL \url{https://arxiv.org/abs/2103.16091}.

\bibitem[Andreas et~al.(2022)Andreas, Martin, Andres, Fisher, Radu, and
  Luc]{Andreas:RepaintDiffusion22}
L.~Andreas, D.~Martin, R.~Andres, Y.~Fisher, T.~Radu, and V.G. Luc.
\newblock Repaint: Inpainting using denoising diffusion probabilistic models.
\newblock In \emph{Proceedings of the IEEE/CVF Conference on Computer Vision
  and Pattern Recognition (CVPR)}, pages 11461--11471, June 2022.

\end{thebibliography}

\end{document}